\pdfoutput=1
\documentclass[journal]{IEEEtran}

\usepackage{amssymb}

\usepackage{graphicx}
\usepackage{tabularx}

\usepackage{algorithm2e}

\usepackage{algpseudocode}

\usepackage{amsmath} 
\usepackage{mathtools}
\usepackage{bm}
\usepackage{esvect}

\usepackage{amsmath}
\usepackage{amssymb}
\usepackage{dsfont}

\usepackage{graphicx}

\usepackage{floatrow}

\usepackage{scalerel}

\usepackage{float}

\usepackage{enumitem}

\SetLabelAlign{Center}{\hfil#1\hfil}
\SetLabelAlign{CenterWithParen}{\hfil(\makebox[0.5em]{#1})\hfil}

\usepackage{caption}
\usepackage{subcaption}

\usepackage{cuted}

\usepackage{graphicx}
\usepackage{dcolumn}
\usepackage{bm}

\usepackage{empheq}
\newcommand*\widefbox[1]{\fbox{\hspace{2em}#1\hspace{2em}}}

\makeatletter
\renewcommand*\env@matrix[1][\arraystretch]{%
  \edef\arraystretch{#1}%
  \hskip -\arraycolsep
  \let\@ifnextchar\new@ifnextchar
  \array{*\c@MaxMatrixCols c}}
\makeatother

\SetKwInput{KwInput}{Input}                
\SetKwInput{KwOutput}{Output}              

\usepackage{balance}

\usepackage{multirow}
\usepackage[dvipsnames]{xcolor}

\renewcommand\arraystretch{1.7}

\usepackage[T1]{fontenc}
\usepackage[utf8]{inputenc}
\usepackage{setspace}
\usepackage{lipsum}
\usepackage{booktabs}
\usepackage{rotating}
\usepackage{siunitx}
\usepackage[skip=0.1\baselineskip]{caption}

\definecolor{mygray}{gray}{0.4}

\usepackage{float}
\floatstyle{plaintop}
\restylefloat{table}

\usepackage[tableposition=top]{caption}

\begin{document}

\title{An Inherent Trade-Off in Noisy Neural Communication with Rank-Order Coding}

\author{Ibrahim~Alsolami and Tomoki~Fukai \\ Okinawa Institute of Science and Technology (OIST)\\Onna, Okinawa, Japan 904-0495}

\markboth{ }%
{Shell \MakeLowercase{\textit{et al.}}: Bare Demo of IEEEtran.cls for IEEE Journals}

\maketitle

\begin{abstract}

Rank-order coding, a form of temporal coding, has emerged as a promising scheme to explain the rapid ability of the mammalian brain. Owing to its speed as well as  efficiency, rank-order coding is increasingly gaining interest in diverse research areas beyond neuroscience. However, much uncertainty still exists about the performance of rank-order coding under noise. Herein we show what information rates are fundamentally possible and what trade-offs are at stake. An unexpected finding in this paper is the emergence of a special class of errors that, in a regime, increase with less noise.

\end{abstract}


\IEEEpeerreviewmaketitle

\maketitle
\section{Introduction}
Currently, there is a growing interest in spiking neural networks in physics and engineering~\cite{markovic2020, vinci2023}. Mainly because of their potential to improve power efficiency, which, to a great extent, depends on the coding scheme employed. For spiking neural networks, a variety of coding schemes are available.   Chief among them is rank-order coding; this coding scheme offers a fundamentally different approach for neural information transmission. In rank-order coding, information is encoded in the order of neural spikes; utilizing this degree of freedom can boost communication speeds and efficiency substantially.

Rank-order coding has been proposed as a faster alternative to the traditional rate coding scheme~\cite{vanrullen2005spike}.
While rate coding is the most widely accepted coding scheme in neuroscience, it has a subtle problem: some experimental observations are hard to reconcile with it because it is slow.  For example, primates can respond selectively to the presentation of 3D objects as quickly as $100$--$150$ ms after the onset of a stimulus ~\cite{thorpe2001,thorpe1989}. This response is too fast to be explained with rate coding as it needs, for a reasonable degree of accuracy, to accumulate spikes over periods much longer than $150$ ms. Similarly, humans who were asked to determine whether a briefly flashed picture ($20$ ms) of a natural scene contains a certain category, such as an animal, could accurately ($\sim 89$--$98\%$) detect whether a category is present or not within a few to several hundreds of milliseconds~\cite{thorpe1996speed}. Rate coding seems to hardly harmonize with such rapid processing speed. 

One could, of course, argue that rate coding across a reasonably large number of neurons could provide speed (in terms of bits/sec). In such an approach, we would have $n$ neurons firing in parallel, and one counts the number of spikes generated within a relatively short time window---compared to a long time window if we had a few neurons. For instance, we would need a longer time window to accumulate $n$ spikes from a single neuron than $n$ neurons firing in parallel. This population rate coding scheme can certainly provide speed, but one must contend with the fact that such an approach is inefficient in terms of bits/neurons~\cite{bonilla}. In fact, the efficiency of this approach is upper-bounded by $\frac{\log_2 (n+1)}{n}$ (bits/neuron)---that is, the more neurons, the less the efficiency. Moreover, the rapid speed of visual processing is likely to be accomplished with very few spikes~\cite{thorpe1989}, and this would require a neural coding scheme whereby a few neurons can communicate efficiently.

Rank-order coding can offer both speed and  efficiency. With $n$ neurons, rank-order coding can encode $\log_2 n!$ bits per transmission compared to $\log_2 (n+1)$ bits per transmission for rate coding. The encoding ability of rank-order coding is vast. Take, for instance, a setting with $10$ neurons. With these neurons, rank-order coding can in principle form $10!=3,628,800$ symbols,  i.e., firing orders of neurons (Fig.~\ref{fig:Illustration}). As $n$ increases, the encoding ability of rank-order coding rapidly accelerates. This vast amount of information available in the arrival order of spikes is often forgotten, and studying neural codes that utilize such arrival order could provide clues on how neurons can transmit information rapidly and efficiently across brain regions. 

Converging evidence suggests that the relative timing, or rank, of neuronal firing plays an important part in encoding information. In retinal ganglion cells of salamanders, the rapid transmission of visual scenes is likely accomplished by encoding information in the relative timing of spikes~\cite{gollisch2008rapid}. It was shown later in a population of retinal ganglion cells of mice that the content of a visual stimulus could be accurately inferred from the wave of the first stimulus-evoked spikes, indicating the importance of the relative timing of spikes to encode sensory information~\cite{portelli2016rank}. Analysis of odor-evoked responses of olfactory neurons of Xenopus laevis (African clawed frogs) demonstrated that the rank of spike latencies is a reliable predictor of odor identity~\cite{junek2010olfactory}.

In addition to its biological applicability, rank-order coding is gaining attention in the field of artificial spiking neural networks, which are becoming popular as they hold great potential in energy-efficient computing~\cite{lemaire2022analytical}. It was shown that spiking neural networks with rank-order coding can achieve a high image-classification accuracy with a relatively small number of spikes in multilayer feedforward networks ~\cite{kheradpisheh} as well as in recurrent networks~\cite{yan2021}. Rank-order coding is also finding favor in hardware implementations of spiking neuromorphic processors;  it was demonstrated that rank-order coding can enhance power efficiency~\cite{frenkel20180} and provide a favorable trade-off between energy consumption and classification accuracy~\cite{frenkel2019morphic}. Recently, it was shown that rank-order coding successfully reduced the on-chip inference latency in neuromorphic devices~\cite{wu2022brain}.

Despite increasing interest in the use of rank-order coding, far too little attention has been paid to the effect of noise on its performance. Our goal here is to analytically understand the impact of noise on the performance of rank-order coding, as noise is unavoidable in any physical system. In rank-order coding, noise can cause spikes to be swapped with each other, giving rise to errors. Herein we study how well rank-order coding performs under noise in terms of information rate (bits/sec) and communication efficiency (bits/neuron). Contrary to intuition, reducing noise does not necessarily reduce all types of errors. Moreover, we show that information rate and communication efficiency cannot be simultaneously maximized due to an intrinsic trade-off between them.\\

\section{Methods}
\begin{figure}[t!]
\captionsetup{}
\centering
\includegraphics[scale=1.06]{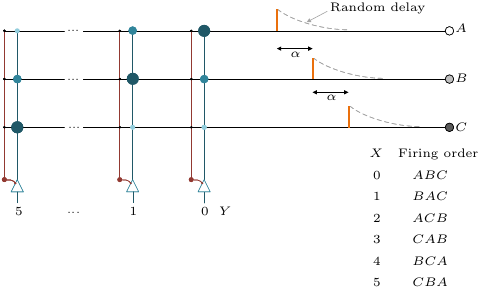}
\caption{Rank-order coding with temporal noise (random delay). Here $\alpha$ is the spacing between  successive spikes before noise is introduced. In this illustration, the magnitude of a synaptic weight is represented by the size of the depicted circles. Here postsynaptic neurons integrate-and-fire and are progressively desensitized by shunting inhibition circuits (red). With shunting inhibition, the sensitivity of a neuron progressively decreases as $\beta^k$, where $k$ is the arrival order of a spike, and $\beta$ is a constant that takes values in the range $0<\beta<1$. A postsynaptic neuron is maximally activated if spikes arrive in the order of its synaptic weights. By setting the firing threshold to this maximum excitation/activation level, a postsynaptic neuron becomes selective to a particular temporal pattern. Due to noise (random delay), an intended spike sequence can be erroneously received. For instance,  in this illustration, noise can cause the sequence ABC ($x=0$) to be erroneously received as CBA ($y= 5$), which impairs both the communication rate (bits/sec) and efficiency (bits/neuron).}
\label{fig:Illustration}
\end{figure}

We consider a noise model in which spike times of presynaptic neurons exhibit random delays characterized by an exponential probability density (Fig.~\ref{fig:Illustration}, Eq.~\ref{eq:pdfFirs1}). Such noise may arise when neurons do not respond instantly to a stimulus, at an expected time, but rather with a random delay (Fig.~\ref {fig:Illustration}).  In rank-order coding, neurons are intensity-to-delay converters: the higher the activation, the earlier a neuron fires. Traditional integrate-and-fire models have this intensity-to-delay property: the higher the membrane potential is, the earlier a neuron will fire. 
Without loss of generality, we hypothesize that a postsynaptic neuron responds selectively and reliably to a particular order of presynaptic spikes. A feed-forward shunting inhibition circuit was suggested as the underlying mechanism of this precise decoding of temporal patterns ~\cite{bonilla}. However, exploring the detailed decoding mechanisms is beyond the scope of this study.

The channel capacity enables us to compute the maximum amount of information postsynaptic neurons can receive and is defined as~\cite{Shannon, Cover}
\begin{equation} \label{eq:Capacity1}
C=\max_{p(x)} I(X;Y)~~~~~\text{(bits/symbol)},
\end{equation}
where $I(X;Y)$ is the mutual information between random variable $X$ (input symbol) and $Y$ (output symbol), and is given by

\begin{equation} \label{eq:Mutual}
I(X;Y)=H(Y)-H(Y|X)~~~~~\text{(bits/symbol)}.
\end{equation}
Here, $H(Y)$ is the entropy of  $Y$, and $H(Y|X)$  is the conditional entropy of $Y$ given $X$.

In Fig.~\ref{fig:Illustration}, we have $(n!)^2$ possible combinations of input and output symbols, where a symbol is defined as a particular order of neural spikes (e.g., the sequence ABC). Here $n$ is the number of presynaptic neurons. The probability of sending symbol $x$ and, because of noise, receiving symbol $y$ is given by the transition probability $p(y|x)$.  The  following probability transition matrix describes such communication channel:

 \begin{equation}\label{eq:Matrix}
\begin{split}
\bm{p(y|x)} =& 
  \left(\begin{smallmatrix}
 p(0|0)& p(1|0)& \cdots & p(n!-1|0)\\
         p(0|1) & p(1|1) & \cdots & p(n!-1|1)\\ 
         \vdots & \vdots & \ddots & \vdots\\ 
         p(0|n!-1) & p(1|n!-1) & \cdots & p(n!-1|n!-1)
  \end{smallmatrix}\right)_{n!\times n!}\\
  \\
  =&\left(\begin{smallmatrix}
         p_0& p_1& \cdots & p_{n!-1}\\
         p_1 & p_0 & \cdots & p_{n!-2}\\ 
         \vdots & \vdots & \ddots & \vdots\\ 
         p_{n!-1} & p_{n!-2} & \cdots & p_{0} 
  \end{smallmatrix}\right)_{n!\times n!}.\\
  \end{split}
\end{equation}

This communication channel is symmetric because rows of the transition matrix are permutations of each other, and so are the columns. The capacity of this channel is achieved by a uniform distribution on the input $X$ ~\cite{Cover}$\left(p(x)=\dfrac{1}{n!}\right)$, which results in a uniform distribution on the output $Y$ $\left(p(y)=\dfrac{1}{n!}\right)$, and is given by 

\begin{equation} \label{eq:Capacity}
\begin{split}
C=& \max_{p(x)} [H(Y)-H(Y|X)]  \\ 
  =& \log_2 n!-H(\bm{r})~~~~~\text{(bits/symbol)},
\end{split}
\end{equation}

\noindent where $H(\bm{r})=-\sum\limits_{j=0}^{n!-1} p_j\log_2 p_j$ is the entropy of a row of matrix $\bm{p(y|x)}$.

\section{Results}
\textbf{Transition probabilities.} Here we determine the transition probabilities  to find the channel capacity in Eq.~(\ref{eq:Capacity}). These probabilities are the likelihood that a particular neural spike sequence is received under the perturbation of noise (random delay, Fig.~\ref{fig:Illustration}). For instance, for three neurons,  $p(CBA|ABC)$ is the probability that the sequence CBA ($y=5$) is erroneously received due to  noise, given that the original noise-free sequence is ABC ($x=0$). It suffices to compute the transition probability of any row of $\bm{p(y|x)}$ because the channel is symmetric~\cite{Cover}. Calculations of the transition probabilities are straightforward but tedious. Therefore, we only evaluate these probabilities when the number of presynaptic neurons is relatively small (see APPENDIX~\ref{Appx1} for derivation).

We can obtain:
\begin{equation} \label{eq:Pr2}
\begin{split}
p_0=p(AB|AB)=& =1-\dfrac{1}{2}e^{-\lambda \alpha} \\
p_1=p(BA|AB)=&=\dfrac{1}{2}e^{-\lambda \alpha}~~,
\end{split}
\end{equation}
for two presynaptic neurons and 
\begin{equation} \label{eq:Pr3}
\begin{split}
&~~~~~p_0=p(ABC|ABC)=1-e^{-\lambda \alpha} + \dfrac{1}{6}e^{-3\lambda \alpha}\\
&~~~~~p_1=p(BAC|ABC)=\dfrac{1}{2}e^{-\lambda \alpha}-\dfrac{1}{2}e^{-2\lambda \alpha} + \dfrac{1}{6}e^{-3\lambda \alpha}\\
&~~~~~p_2=p(ACB|ABC)=\dfrac{1}{2}e^{-\lambda \alpha}-\dfrac{1}{3}e^{-3\lambda \alpha} \\
&~~~~~p_3=p(CAB|ABC)=\dfrac{1}{6}e^{-3\lambda \alpha} \\
&~~~~~p_4=p(BCA|ABC)=\dfrac{1}{2}e^{-2\lambda \alpha} -\dfrac{1}{3}e^{-3\lambda \alpha} \\
&~~~~~p_5=p(CBA|ABC)=\dfrac{1}{6}e^{-3\lambda \alpha}~~,
\end{split}
\end{equation} 
for three presynaptic neurons. In the above expressions, $\alpha$ is the spacing between successive spikes before noise is introduced, and $\lambda$ is the rate parameter of the exponential distribution of the noise. Results with four presynaptic neurons are shown in APPENDIX~\ref{Appx1}.

\begin{figure}[t!]
        \subfloat[]{%
            \includegraphics[scale=.32]{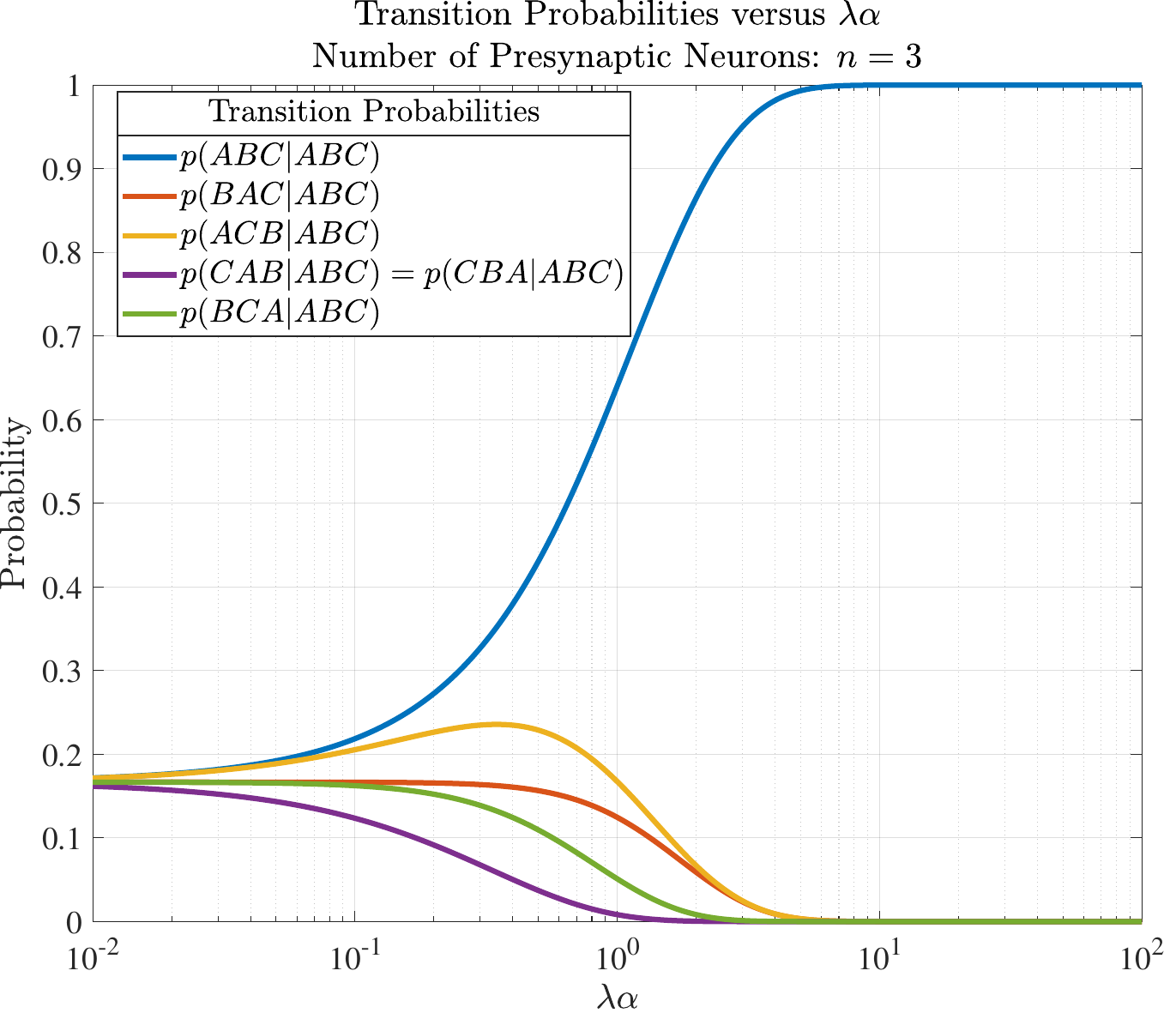}%
            \label{fig:All_Pr}
        }\\
        \subfloat[]{%
            \includegraphics[scale=.35]{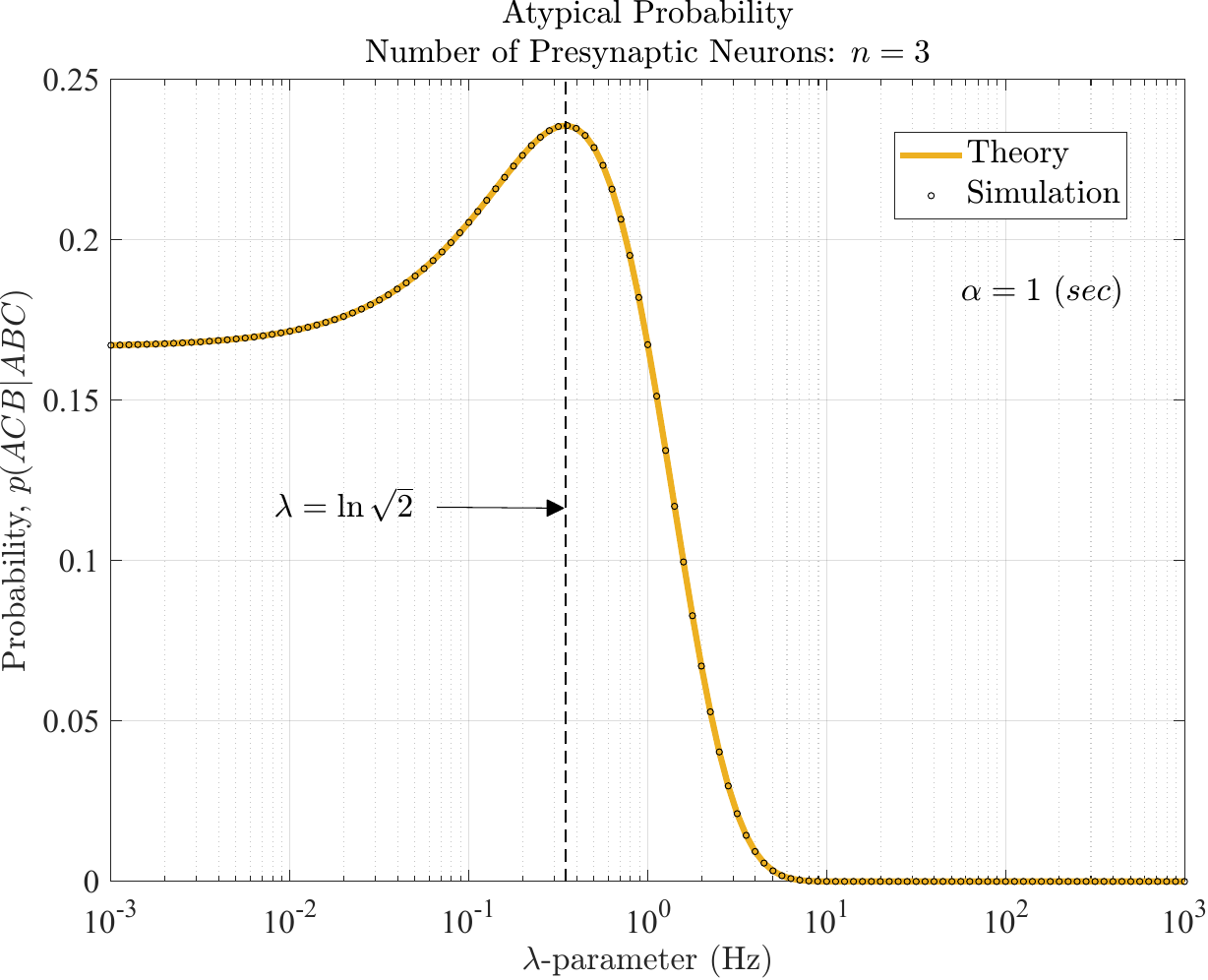}%
            \label{fig:Divergent_Pr}
        }\\
        \subfloat[]{%
            \includegraphics[scale=2.2]{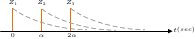}%
            \label{fig:T}%
        }
        \caption{Transition probabilities and spikes of rank-order coding with
temporal noise. (a) Transition probabilities for three presynaptic neurons ($n=3$). (b) Atypical probability $p(ACB|ABC)$; this  probability reflects the likelihood that noise causes the sequence $ABC$ to be erroneously received as $ACB$. In the range $0<\lambda<\ln\sqrt{2}$,  the probability of an error increases with  $\lambda$ (the higher the value of $\lambda$, the less the noise). Here $\alpha$ is arbitrarily set to 1 (sec), and  the number of samples per point used in the simulation is $10^9$. A similar phenomenon, where errors momentarily increase, can also be observed by fixing $\lambda$ and varying $\alpha$; that is, errors increase as the spacing between neural spikes, $\alpha$, increases. (c) Spikes of rank-order coding with temporal noise (random delay).  When $Z_2>Z_3$, more space for $Z_1$ has been made to take the position of the smallest value in the interval $(0,2\alpha)$; this causes the probability of the event $ACB$ ($Z_1<Z_3<Z_2$) to increase.}
        \label{fig:Expl}
\end{figure}

\begin{figure*}[tbh!]
\centering
   \includegraphics[scale=0.55]{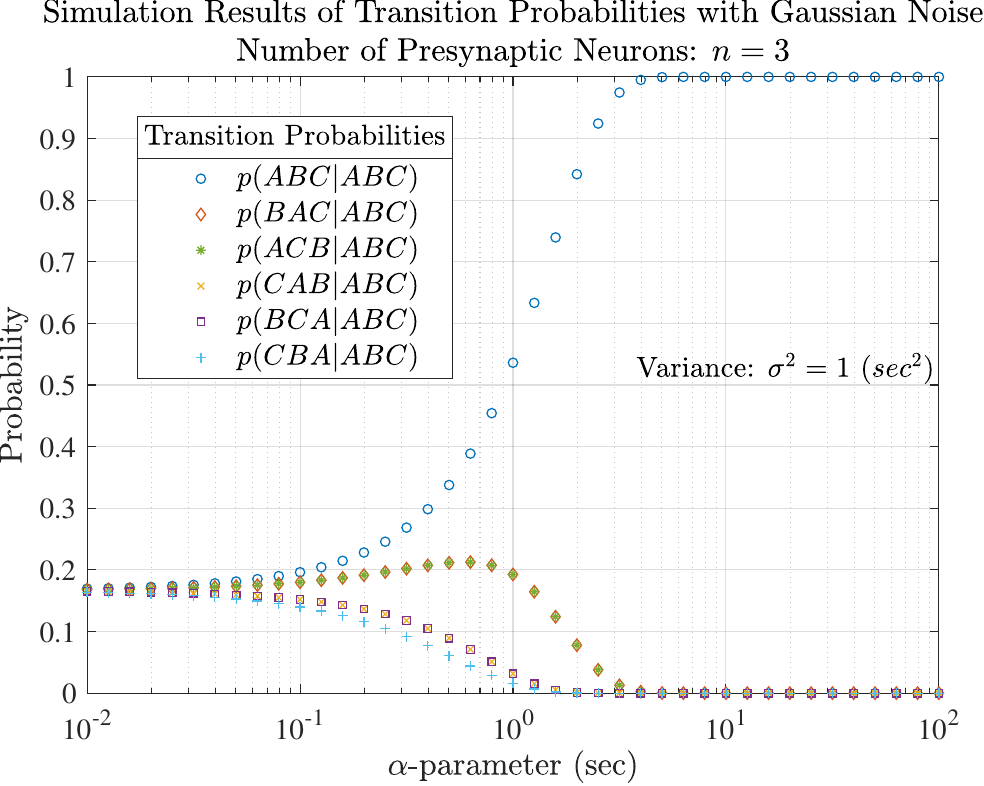}
   \caption{Simulation results of transition probabilities for three presynaptic neurons ($n=3$) with Gaussian noise. Here $p(y|x)$ is the probability of sending sequence $x$ and receiving sequence $y$. The variance of the Gaussian distribution, $\sigma^2$, is arbitrarily set to 1 (sec$^2$). Additionally, $\alpha$ is the original spacing between successive spikes before noise is introduced. The number of samples per point used in the simulation is $10^9$.}
   \label{fig:Pr3G}
\end{figure*}

\begin{figure*}[tbh!]
\centering
   \includegraphics[scale=.8]{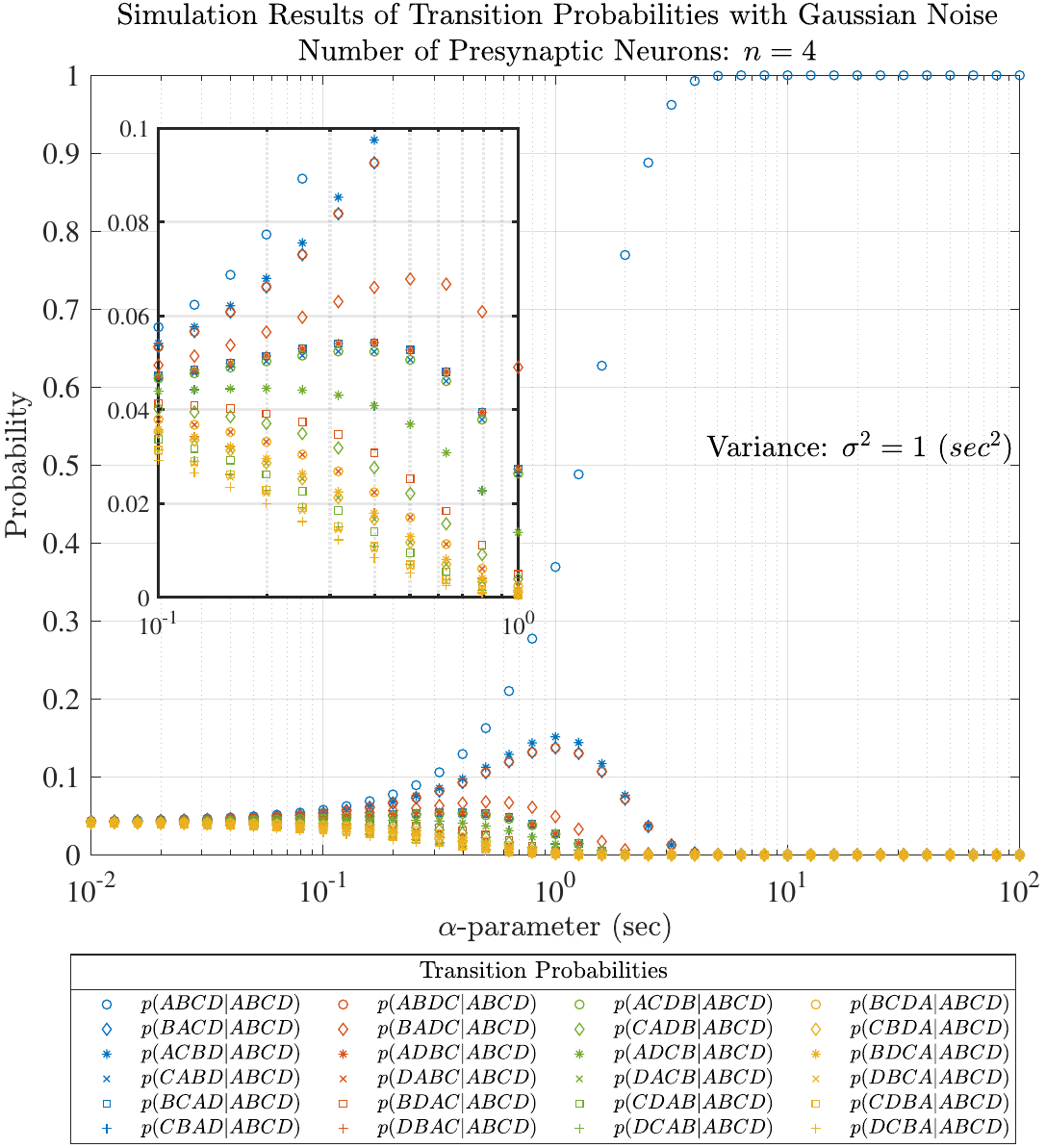}
   \caption{Simulation results of transition probabilities for four presynaptic neurons ($n=4$) with Gaussian noise. Here $p(y|x)$ is the probability of sending sequence $x$ and receiving sequence $y$. The variance of the Gaussian distribution, $\sigma^2$, is arbitrarily set to 1 (sec$^2$). Additionally, $\alpha$ is the original spacing between successive spikes before noise is introduced. The number of samples per point used in the simulation is $10^9$.}
   \label{fig:Pr4G}
\end{figure*}

As expected, when $\lambda\alpha$ increases, error probabilities decrease (Fig.~\ref{fig:All_Pr}). There is a notable exception, however. In the range $0\le\lambda\alpha\le \ln\sqrt{2}$, the error probability $p(ACB|ABC)$ increases. This can be viewed in two different ways: \textbf{1)} For a fixed value of $\alpha$, as the noise decreases (that is, $\lambda$ increases), the probability of this type of error increases (Fig.~\ref{fig:Divergent_Pr}). \textbf{2)} For a fixed value of $\lambda$, as the spacing between spikes ($\alpha$) increases the error probability increases as well. A similar phenomenon is also observed when we have four ($n=4$) presynaptic neurons (see Fig.~\ref{fig:APPXPr4}). Namely, the following error probabilities:
\begin{eqnarray*}\label{eq:Para}
&&p(ABDC|ABCD),~p(ACBD|ABCD),\\
&&p(ACDB|ABCD),~p(ADBC|ABCD),\\
&&p(ADCB|ABCD),~p(BACD|ABCD),\\
&&p(BADC|ABCD),~p(BCAD|ABCD),\\
&&p(BCDA|ABCD),~p(BDAC|ABCD),
\end{eqnarray*}
and $p(BDCA |ABCD)$. These error probabilities momentarily increase with less noise, which is counter-intuitive: errors typically decrease with less noise---not the opposite. Throughout this study, we shall refer to this class of probabilities as \textit{atypical probabilities}.  This type of error is not limited to exponential noise; it can also be observed, for example,  with Gaussian noise (see Figs.~\ref{fig:Pr3G} and \ref{fig:Pr4G}).

\textbf{Why do errors increase when we have less noise?} The emergence of atypical probabilities can be explained as follows. Let the probability $P(ACB|ABC)$  serve as an example  (Fig.~\ref{fig:Expl}). Moreover, let random variables $Z_i$ ($i=$1, 2, 3) represent a spike's latency after the perturbation of noise; here $i$ denotes the index of the $i^\text{th}$ presynaptic neuron (See APPENDIX for notation details). For the event $Z_1<Z_3<Z_2$, or equivalently the sequence $ACB$, to occur, the following two conditions should be simultaneously satisfied: (i) Random variable $Z_2$ needs to be the largest value  and (ii) random variable $Z_1$ needs to be the smallest value. 

\begin{enumerate}[label=\roman*,align=CenterWithParen]
 \item The probability of $Z_2$ being the largest value (i.e., $Z_2>Z_3$) decreases with $\lambda$ (the larger the value of $\lambda$, the less the noise) because the amount of overlap between the distributions of $Z_2$ and $Z_3$ decreases as $\lambda$ increases.

 \item In contrast, the probability of $Z_1$ being the smallest value increases with $\lambda$ in the interval $(0,2\alpha)$ for the following reason. The event $Z_2>Z_3$ implies that $ Z_2 >2\alpha$~(Fig.~\ref{fig:T} and Eq.~\ref{eq:pdfFirs1}). Thus, when $Z_2>Z_3$, more space (from $\alpha$ to $ 2\alpha$) for $Z_1$ has been made to take the position of the smallest value in the interval $(0,2\alpha)$, thereby increasing the likelihood of the neural order $ACB$ ($Z_1<Z_3<Z_2$).
\end{enumerate}
Factor (i) causes $P(ACB|ABC)$ to decrease, whereas factor (ii) causes $P(ACB|ABC)$ to increase. The net effect of factors (i) and (ii) is  $P(Z_1<2\alpha< Z_3<Z_2)$, which is a concave function. This component brings about a rare regime in which errors increase with $\lambda$ (or, equivalently, with $\alpha$). Mathematically, the  probability $P(Z_1<2\alpha<Z_3<Z_2)$ can be obtained by splitting the integration region of $P(ACB|ABC)$ into two parts:
\vspace{-1.05cm}
\begin{strip}
\begin{align} \label{eq:Explain}
& P(ACB|ABC)= P(Z_1<Z_3<Z_2~|~ABC)=\int_{2\alpha}^{\infty} \int_{2\alpha}^{z_2} \int_{0}^{z_3}      f_1(z_1) f_3(z_3) f_2(z_2)  ~\,dz_1\, dz_3\, dz_2  \\ 
&= \underbrace{\left(\int_{2\alpha}^{\infty} \int_{2\alpha}^{z_2} \int_{0}^{2\alpha}      f_1(z_1) f_3(z_3) f_2(z_2)  ~\,dz_1\, dz_3\, dz_2\right)}_\text{Concave: $P(Z_1<2\alpha<Z_3<Z_2)$}+\underbrace{\left(\int_{2\alpha}^{\infty} \int_{2\alpha}^{z_2} \int_{2\alpha}^{z_3}      f_1(z_1) f_3(z_3) f_2(z_2)  ~\,dz_1\, dz_3\, dz_2\right)}_\text{Convex: $P(2\alpha < Z_1<Z_3<Z_2)$} \nonumber \\ 
&= \underbrace{\left(  \underbrace{\dfrac{1}{2}e^{-\lambda \alpha}}_\text{Factor (i): $P(2\alpha<Z_3<Z_2)$}\times~~~ \underbrace{\left( 1-e^{-2\lambda \alpha} \right)}_\text{Factor (ii): $P(Z_1<2\alpha)$} \right)}_\text{Concave: $P(Z_1<2\alpha<Z_3<Z_2)=P(Z_1<2\alpha)~\times ~ P(2\alpha<Z_3<Z_2)$}~~~~~~~~~~~+ \underbrace{\left( \dfrac{1}{6}e^{-3\lambda \alpha}    \right)}_\text{Convex: $P(2\alpha<Z_1<Z_3<Z_2)$} =  \dfrac{1}{2}e^{-\lambda \alpha}-\dfrac{1}{3}e^{-3\lambda \alpha}~~\nonumber .
\end{align}
\end{strip}

\noindent Here $f_i(z)$ is the  probability density function (\textit{pdf}) of the $i^{\text{th}}$ presynaptic neuron, $i\in\{1,2,\dots,n\}$. This \textit{pdf} describes the likelihood of observing a randomly delayed spike, by noise, at time $z$ and is given by
\begin{equation} \label{eq:pdfFirs1}
f_i(z)=
\begin{cases}
\lambda~e^{-\lambda \bigl (z-(i-1)\alpha\bigr)}&,~\text{if}~z\geq(i-1)\alpha\\
0&,~\text{otherwise}~~~.
\end{cases}
\end{equation}


\begin{figure}[t!]
        \subfloat[]{%
            \includegraphics[scale=.405]{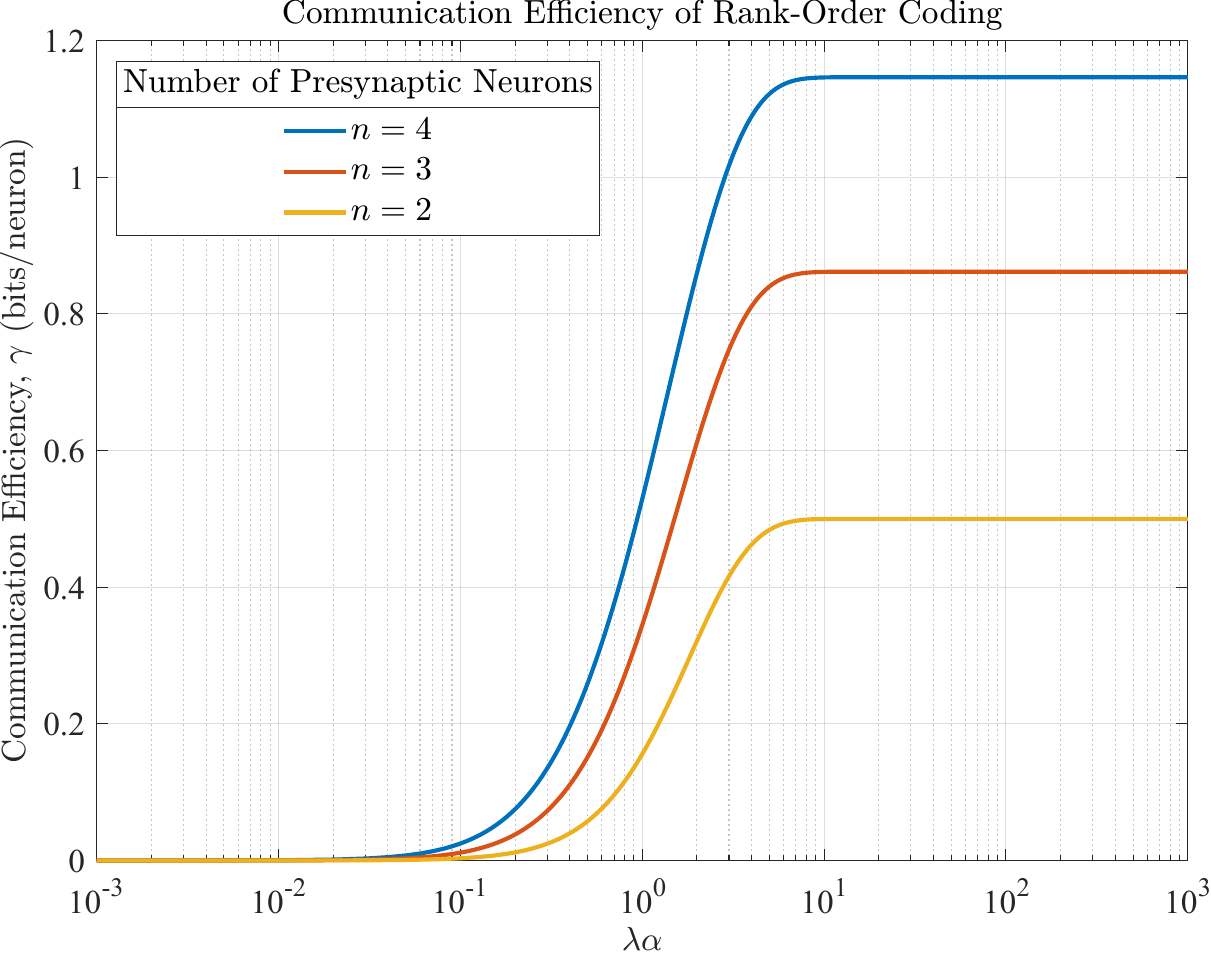}%
            \label{fig:E}%
        }\\
        \subfloat[]{%
            \includegraphics[scale=.405]{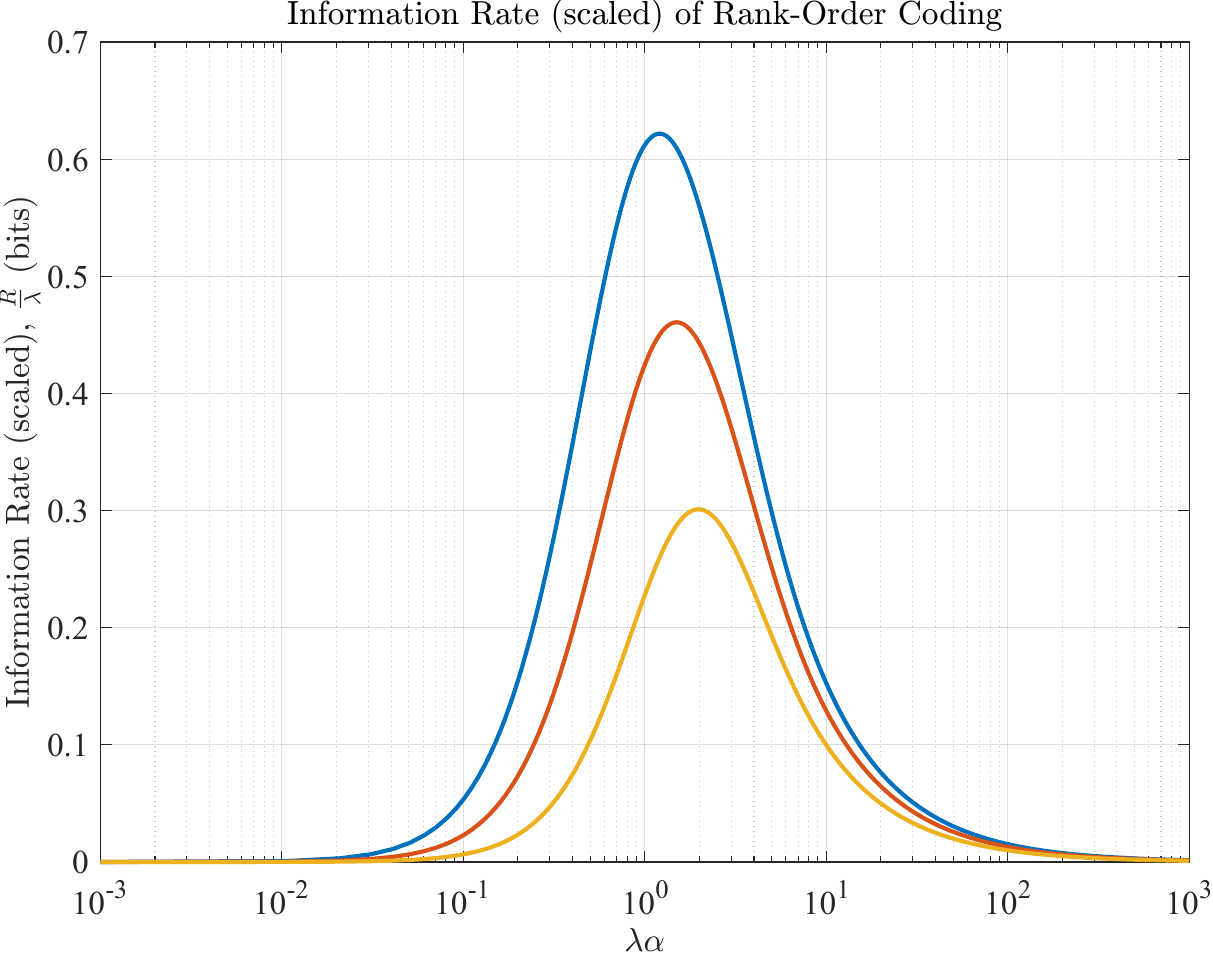}%
            \label{fig:R}%
        }\\
        \subfloat[]{%
            \includegraphics[scale=.3705]{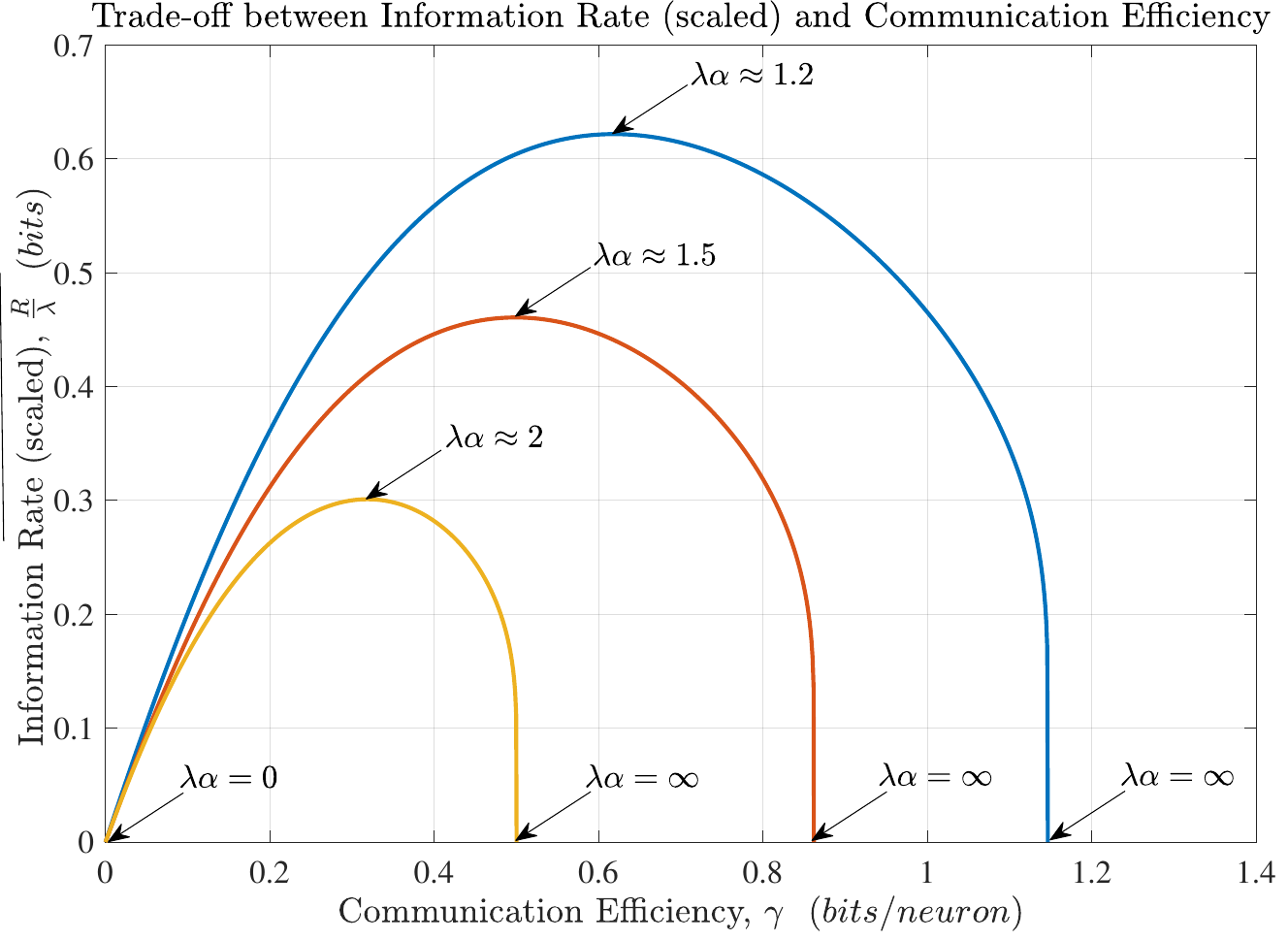}%
            \label{fig:Trade_Off}%
        }
        \caption{Performance of rank-order coding.   (a) Communication efficiency.  (b) Information rates. Here we plot the scaled version ($\frac{R}{\lambda}$) of $R$ rather than $R$ as it eliminates the need to display $R$ for various combinations of $\lambda$ and $\alpha$. (c) The trade-off between information rates and efficiency. }
        \label{fig:fig}
\end{figure}

\noindent\textbf{Communication efficiency and information rate.} Figure~\ref{fig:E} shows the performance of rank-order coding in terms of communication efficiency, which is defined as
\begin{equation}\label{eq:Info_Efficiency}
   \gamma=\frac{C}{n}~~~~~~~\text{(bits/neuron)},
\end{equation}

\noindent where $C$ can be calculated by using Eq.~\ref{eq:Capacity}. The communication efficiency increases monotonically with $\lambda\alpha$. This increase in efficiency eventually plateaus and is asymptotically bounded by $\gamma^*=\lim_{\lambda\alpha\to\infty}\frac{C}{n}=\frac{\log_2(n!)}{n}$ (bits/neuron). Moreover, the higher the value of $n$, the more efficient the communication is.

We can further evaluate the performance of rank-order coding in terms of information rate, which is defined as
\begin{equation}\label{eq:Info_Rate}
   R=\frac{C}{\bar{T}}~~~~~~~\text{(bits/sec)}.
\end{equation}
 \noindent In the absence of noise, the average symbol duration $\bar{T}$ (that is, the time difference between the first and last spikes of a rank-order coding symbol) is $(n-1)\alpha$. However, with noise, the average symbol duration increases and is given by
\begin{equation}\label{eq:SymbolDuration2}
\bar{T}=\alpha+\frac{1}{\lambda}e^{-\lambda \alpha}~~~\text{(sec/symbol)}
\end{equation}
for two presynaptic  neurons ($n=2$),
\begin{equation}\label{eq:SymbolDuration3}
\bar{T}=2\alpha+\frac{1}{\lambda}e^{-\lambda \alpha}+\frac{1}{2\lambda}e^{-2\lambda \alpha}~~~\text{(sec/symbol)}
\end{equation}
for three presynaptic  neurons ($n=3$), and
\begin{align}\label{eq:SymbolDuration4}
\bar{T}= 3\alpha&+\frac{1}{\lambda}e^{-\lambda \alpha}+\frac{1}{2\lambda}e^{-2\lambda \alpha}+\frac{1}{2\lambda}e^{-3\lambda \alpha}-\frac{1}{6\lambda}e^{-4\lambda \alpha} \nonumber \\
& -\frac{1}{6\lambda}e^{-5\lambda \alpha}+\frac{1}{6\lambda}e^{-6\lambda \alpha}~~~\text{(sec/symbol)}
\end{align}
for four presynaptic neurons ($n=4$) (see APPENDIX~\ref{Appx3} for  derivation). In Fig.~\ref{fig:R}, we display the (scaled) information rate as a function of $\lambda \alpha$. The information rate is a non-monotonic function of $\lambda \alpha$ and increases with $n$. Moreover, there is an optimal operating point  at which the information rate is maximized; beyond this critical point, the information rate rapidly diminishes.

In a noisy environment, there exists an inherent trade-off between the communication efficiency of rank-order coding and its information rate. The communication efficiency continuously increases with $\lambda \alpha$ (Fig.~\ref{fig:E}), but this gain in efficiency comes at the cost of a loss in the information rate once $\lambda\alpha$ is beyond a critical value (Fig.~\ref{fig:R}). A range of trade-offs is shown in Fig.~\ref{fig:Trade_Off}, in which the value of $\lambda \alpha$ is varied and the pair $ (\gamma, \frac{R}{\lambda})$ is displayed. The resultant curves represent upper-bounds of achievable information rates and communication efficiencies. Parameter $\alpha$ provides a means to control the trade-off between information rate and efficiency.

\section{Discussion}
This paper set out to study the impact of noise on the performance of rank-order coding. Rank-order coding is advantageous as it utilizes the order of neural spikes, enabling it to boost communication speeds. A disadvantage, however, is that it is susceptible to temporal noise, which can swap presynaptic spikes with each other causing errors at postsynaptic neurons. As such, we considered noise in the form of a random delay to gain insights into the performance of rank-order coding in terms of information rate and communication efficiency.

In  noisy environments,  the information rate and communication efficiency  depend at least on three factors: the spacing between spikes $~\alpha$, the rate parameter $~\lambda$, and the number of presynaptic neurons $~n$. The higher the value of $\lambda \alpha$, the more efficient the communication is. However, increasing $\lambda \alpha$ beyond an optimal operating point has the adverse effect of reducing the information rate.  Additionally, we found a class of error probabilities that increase with less noise. This result is counter-intuitive because errors commonly decrease with less noise---not the opposite. The presence of such error probabilities raises a need for special care in designing error correction schemes for neuromorphic devices that employ rank-order coding.

We revealed that rank-order coding has an inherent trade-off between information rate and communication efficiency. This result could provide insights to better understand what trade-offs neurons in different brain regions make (under the rank-order coding hypothesis) between the conflicting needs to be fast and, at the same time, efficient. For example, it is likely that neurons at the early stage of the visual-processing pathway (e.g., the retina) prioritize speed over efficiency. However, efficiency may be favored over speed at later stages as information is likely at/near its final (decision-making) destination. The trade-off result also offers a realistic picture of neuromorphic computing with rank-order coding: information rate and communication efficiency cannot be simultaneously maximized---a compromising trade-off between them needs to be made (Fig.~\ref{fig:Trade_Off}).

In the present study, we assumed that postsynaptic neurons respond selectively to a particular order of spikes (temporal pattern). Studies have shown that cortical neurons exhibit such selectivity to temporal input sequences~\cite{Branco2010}. Various biological mechanisms of temporal pattern detection have been proposed (e.g., ~\cite{gutig2006}). A feed-forward shunting inhibition circuit, which progressively desensitizes a postsynaptic neuron as spikes arrive (see Fig.~\ref{fig:Illustration}), may  accomplish selectivity to a particular temporal pattern~\cite{bonilla}. In such a setting, a postsynaptic neuron would be maximally activated (and fire only) if spikes arrive in the order of its synaptic weights. A small portion of extremely strong synapses observed in log-normally distributed synaptic weights~\cite{Buzsaki2014} may enhance this progressive desensitization. 

Generalization of our results to an arbitrary number of neurons is a promising research extension. Another avenue of extension would be to assess the performance of rank-order coding when not all spikes arrive at postsynaptic neurons; some presynaptic neurons, for instance, may misfire. The combination of such noise with temporal noise can significantly affect performance. Nonetheless, hybrid noise would only affect quantitative results but would not change the qualitative results of this study.

To conclude, rank-order coding can provide speed and efficiency, but noise imposes a trade-off between them. The results of this study offer a novel insight into the performance of rank-order coding.

\vspace{.7cm}

\section*{Acknowledgment}
The authors would like to thank Balashwethan Chockalingam and Thomas Burns for their suggestions and comments. We are grateful for the help and support provided by the Scientific Computing and Data Analysis section of the Research Support Division at OIST. T.F. acknowledges support from KAKENHI grants JP19H04994 and JP18H05213.

\balance
\bibliographystyle{IEEEtran}
\bibliography{My.bib}


\appendices
\onecolumn
\section{Transition Probabilities of Rank-Order Coding with Exponential Noise (Theory)}\label{Appx1}

\textbf{Two Neurons.} For two presynaptic neurons, the transition probabilities of the first row of Eq.~(3) are

\begin{equation*} \label{eq:APPXPr2}
\begin{split}
p_0=p(AB|AB)=&\int_{\alpha}^{\infty} \int_{0}^{z_2} f_1(z_1) f_2(z_2)\,dz_1\, dz_2 =1-\dfrac{1}{2}e^{-\lambda \alpha} \\
p_1=p(BA|AB)=& \int_{\alpha}^{\infty} \int_{\alpha}^{z_1} f_2(z_2) f_1(z_1)\,dz_2\, dz_1 =\dfrac{1}{2}e^{-\lambda \alpha}~~.
\end{split}
\end{equation*}

\noindent Let $Z_i$ be a random variable and  $f_i(z)$ its  probability density function (\textit{pdf}); here $f_i(z)$ is the \textit{pdf} of the $i^{\text{th}}$ presynaptic neuron, $i\in\{1,2,\dots,n\}$. This \textit{pdf} describes the likelihood of observing a randomly delayed spike, by noise, at time $z$ and is given by

\begin{equation} \label{eq:pdfFirs2}
f_i(z)=
\begin{cases}
\lambda~e^{-\lambda \bigl (z-(i-1)\alpha\bigr)}&,~\text{if}~z\geq(i-1)\alpha\\
0&,~\text{otherwise}
\end{cases}
\end{equation}

\noindent here $\alpha$ is the spacing between successive spikes before noise is introduced, and $\lambda$ is the rate parameter of the exponential distribution of the noise. Herein random variables, $Z_i$, are independent.\\

\begin{figure*}[tbh!]
\centering
   \includegraphics[scale=.35]{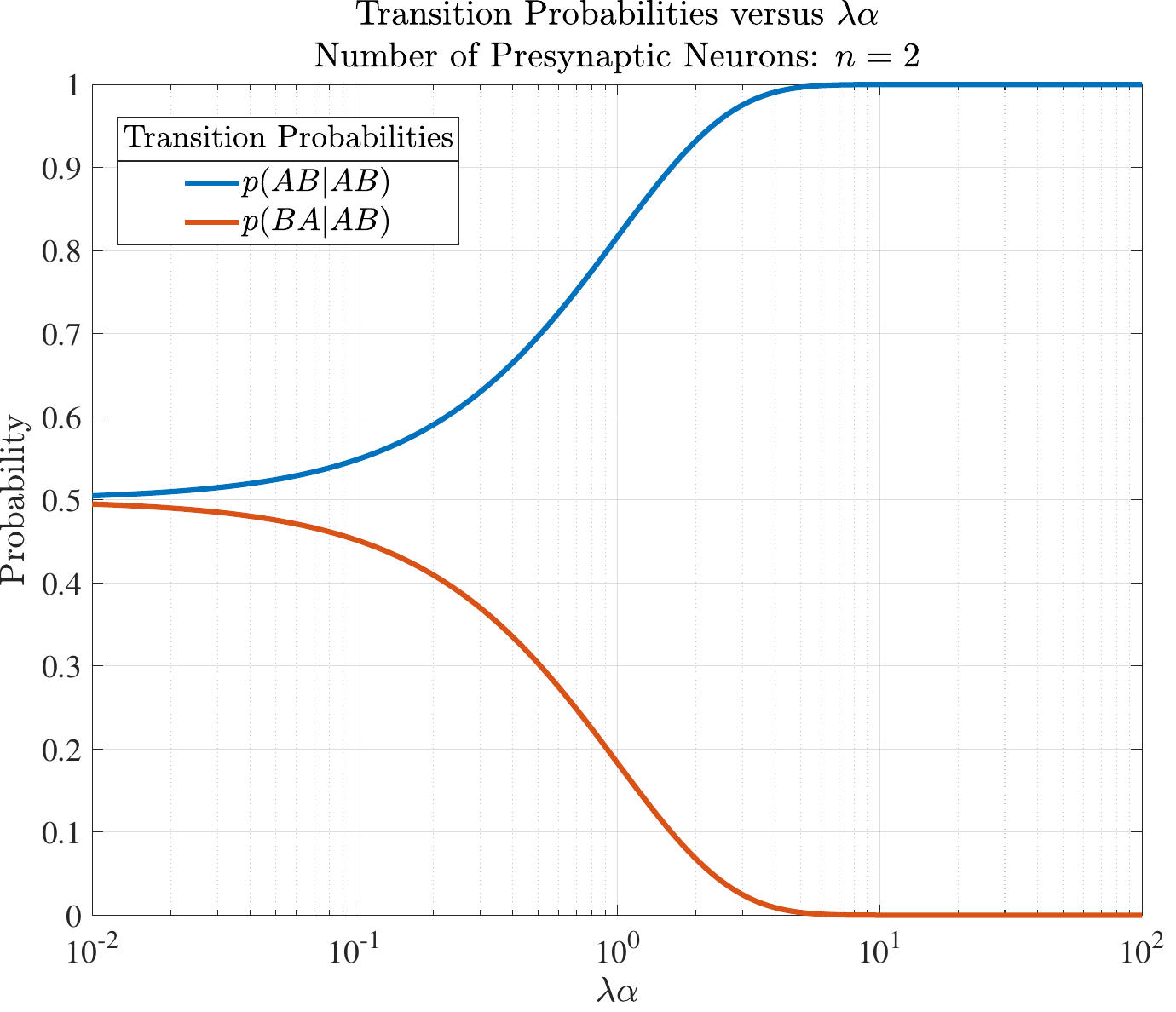}
   \caption{Transition probabilities for two presynaptic neurons ($n=2$) with exponential noise. Here $p(y|x)$ is the probability of sending sequence $x$ and receiving sequence $y$.}
   \label{fig:APPXPr2}
\end{figure*}

\noindent Fig.~\ref{fig:APPXPr2} shows the transition probabilities for two presynaptic neurons ($n=2$).

\noindent\textbf{Three Neurons.} For three presynaptic neurons, the transition probabilities of the first row of Eq.~(3) are

\begin{equation*} \label{eq:APPXPr3}
\begin{split}
p_0=p(ABC|ABC)=&\int_{2\alpha}^{\infty} \int_{\alpha}^{z_3} \int_{0}^{z_2}          f_1(z_1) f_2(z_2) f_3(z_3)  ~\,dz_1\, dz_2\, dz_3 =1-e^{-\lambda \alpha} + \dfrac{1}{6}e^{-3\lambda \alpha}\\
p_1=p(BAC|ABC)=&\int_{2\alpha}^{\infty} \int_{\alpha}^{z_3} \int_{\alpha}^{z_1}     f_2(z_2) f_1(z_1) f_3(z_3)  ~\,dz_2\, dz_1\, d_3 =\dfrac{1}{2}e^{-\lambda \alpha}-\dfrac{1}{2}e^{-2\lambda \alpha} + \dfrac{1}{6}e^{-3\lambda \alpha}\\
p_2=p(ACB|ABC)=&\int_{2\alpha}^{\infty} \int_{2\alpha}^{z_2} \int_{0}^{z_3}         f_1(z_1) f_3(z_3) f_2(z_2)  ~\,dz_1\, dz_3\, dz_2 =\dfrac{1}{2}e^{-\lambda \alpha}-\dfrac{1}{3}e^{-3\lambda \alpha} \\
p_3=p(CAB|ABC)=&\int_{2\alpha}^{\infty} \int_{2\alpha}^{z_2} \int_{2\alpha}^{z_1}   f_3(z_3) f_1(z_1) f_2(z_2) ~\,dz_3\, dz_1\, dz_2 =\dfrac{1}{6}e^{-3\lambda \alpha} \\
p_4=p(BCA|ABC)=&\int_{2\alpha}^{\infty} \int_{2\alpha}^{z_1} \int_{\alpha}^{z_3}    f_2(z_2) f_3(z_3) f_1(z_1)  ~\,dz_2\, dz_3\, dz_1 =\dfrac{1}{2}e^{-2\lambda \alpha} -\dfrac{1}{3}e^{-3\lambda \alpha} \\
p_5=p(CBA|ABC)=&\int_{2\alpha}^{\infty} \int_{2\alpha}^{z_1} \int_{2\alpha}^{z_2}   f_3(z_3) f_2(z_2) f_1(z_1)  ~\,dz_3\, dz_2\, dz_1 =\dfrac{1}{6}e^{-3\lambda \alpha}~~~.\\
\end{split}
\end{equation*}

Fig.~\ref{fig:All_Pr} shows the transition probabilities for three presynaptic neurons ($n=3$).

\newpage
\noindent\textbf{Four Neurons.} For four presynaptic  neurons, the transition probabilities of the first row of Eq.~(3) are

{ \footnotesize %
\begin{equation} 
\begin{split}\label{eq:APPXPr4}
p_0=p(ABCD|ABCD)=&\int_{3\alpha}^{\infty} \int_{2\alpha}^{z_4} \int_{\alpha}^{z_3} \int_{0}^{z_2}         f_1(z_1) f_2(z_2) f_3(z_3) f_4(z_4)~\,dz_1\,dz_2\, dz_3\, dz_4 =1-\dfrac{3}{2}e^{-\lambda \alpha} + \dfrac{1}{4}e^{-2\lambda \alpha} +\dfrac{1}{3}e^{-3\lambda \alpha}-\dfrac{1}{24}e^{-6\lambda \alpha}\nonumber\\
p_1=p(BACD|ABCD)=&\int_{3\alpha}^{\infty} \int_{2\alpha}^{z_4} \int_{\alpha}^{z_3} \int_{\alpha}^{z_1}    f_2(z_2) f_1(z_1) f_3(z_3) f_4(z_4)~\,dz_2\,dz_1\, dz_3\, dz_4 =\dfrac{1}{2}e^{-\lambda \alpha}-\dfrac{3}{4}e^{-2\lambda \alpha}+\dfrac{1}{6}e^{-3\lambda \alpha}+\dfrac{1}{6}e^{-4\lambda \alpha}-\dfrac{1}{24}e^{-6\lambda \alpha} \nonumber\\
p_2=p(ACBD|ABCD)=&\int_{3\alpha}^{\infty} \int_{2\alpha}^{z_4} \int_{2\alpha}^{z_2} \int_{0}^{z_3}        f_1(z_1) f_3(z_3) f_2(z_2) f_4(z_4)~\,dz_1\,dz_3\, dz_2\, dz_4=\dfrac{1}{2}e^{-\lambda \alpha}-\dfrac{1}{2}e^{-2\lambda \alpha}-\dfrac{1}{6}e^{-3\lambda \alpha}+\dfrac{1}{4}e^{-4\lambda \alpha}-\dfrac{1}{24}e^{-6\lambda \alpha} \nonumber\\
p_3=p(CABD|ABCD)=&\int_{3\alpha}^{\infty} \int_{2\alpha}^{z_4} \int_{2\alpha}^{z_2} \int_{2\alpha}^{z_1}  f_3(z_3) f_1(z_1) f_2(z_2) f_4(z_4)~\,dz_3\,dz_1\, dz_2\, dz_4=\dfrac{1}{6}e^{-3\lambda \alpha}-\dfrac{1}{4}e^{-4\lambda \alpha}+\dfrac{1}{6}e^{-5\lambda \alpha}-\dfrac{1}{24}e^{-6\lambda \alpha} \nonumber\\
p_4=p(BCAD|ABCD)=&\int_{3\alpha}^{\infty} \int_{2\alpha}^{z_4} \int_{2\alpha}^{z_1} \int_{\alpha}^{z_3}   f_2(z_2) f_3(z_3) f_1(z_1) f_4(z_4)~\,dz_2\,dz_3\, dz_1\, dz_4=\dfrac{1}{2}e^{-2\lambda \alpha}-\dfrac{5}{6}e^{-3\lambda \alpha}+\dfrac{5}{12}e^{-4\lambda \alpha}-\dfrac{1}{24}e^{-6\lambda \alpha} \nonumber\\
p_5=p(CBAD|ABCD)=&\int_{3\alpha}^{\infty} \int_{2\alpha}^{z_4} \int_{2\alpha}^{z_1} \int_{2\alpha}^{z_2}  f_3(z_3) f_2(z_2) f_1(z_1) f_4(z_4)~\,dz_3\,dz_2\, dz_1\, dz_4= \dfrac{1}{6}e^{-3\lambda \alpha}-\dfrac{1}{4}e^{-4\lambda \alpha}+\dfrac{1}{6}e^{-5\lambda \alpha}-\dfrac{1}{24}e^{-6\lambda \alpha} \nonumber\\
p_6=p(ABDC|ABCD)=&\int_{3\alpha}^{\infty} \int_{3\alpha}^{z_3} \int_{\alpha}^{z_4} \int_{0}^{z_2}            f_1(z_1) f_2(z_2) f_4(z_4) f_3(z_3)~\,dz_1\,dz_2\, dz_4\, dz_3 =\dfrac{1}{2}e^{-\lambda \alpha}-\dfrac{1}{4}e^{-2\lambda \alpha}-\dfrac{1}{3}e^{-3\lambda \alpha}+\dfrac{1}{8}e^{-6\lambda \alpha} \nonumber\\
p_7=p(BADC|ABCD)=&\int_{3\alpha}^{\infty} \int_{3\alpha}^{z_3} \int_{\alpha}^{z_4} \int_{\alpha}^{z_1}       f_2(z_2) f_1(z_1) f_4(z_4) f_3(z_3)~\,dz_2\,dz_1\, dz_4\, dz_3  =\dfrac{1}{4}e^{-2\lambda \alpha} -\dfrac{1}{3}e^{-4\lambda \alpha}+\dfrac{1}{8}e^{-6\lambda \alpha} \nonumber\\
p_8=p(ADBC|ABCD)=&\int_{3\alpha}^{\infty} \int_{3\alpha}^{z_3} \int_{3\alpha}^{z_2} \int_{0}^{z_4}           f_1(z_1) f_4(z_4) f_2(z_2) f_3(z_3)~\,dz_1\,dz_4\, dz_2\, dz_3  =\dfrac{1}{6}e^{-3\lambda \alpha} - \dfrac{1}{8}e^{-6\lambda \alpha} \nonumber\\
p_9=p(DABC|ABCD)=&\int_{3\alpha}^{\infty} \int_{3\alpha}^{z_3} \int_{3\alpha}^{z_2} \int_{3\alpha}^{z_1}     f_4(z_4) f_1(z_1) f_2(z_2) f_3(z_3)~\,dz_4\,dz_1\, dz_2\, dz_3 = \dfrac{1}{24}e^{-6\lambda \alpha}\nonumber\\
p_{10}=p(BDAC|ABCD)=&\int_{3\alpha}^{\infty} \int_{3\alpha}^{z_3} \int_{3\alpha}^{z_1} \int_{\alpha}^{z_4}  f_2(z_2) f_4(z_4) f_1(z_1) f_3(z_3)~\,dz_2\,dz_4\, dz_1\, dz_3 =\dfrac{1}{6}e^{-4\lambda \alpha} - \dfrac{1}{8}e^{-6\lambda \alpha} \nonumber\\
p_{11}=p(DBAC|ABCD)=&\int_{3\alpha}^{\infty} \int_{3\alpha}^{z_3} \int_{3\alpha}^{z_1} \int_{3\alpha}^{z_2}  f_4(z_4) f_2(z_2) f_1(z_1) f_3(z_3)~\,dz_4\,dz_2\, dz_1\, dz_3  =\dfrac{1}{24}e^{-6\lambda \alpha}\\
p_{12}=p(ACDB|ABCD)=&\int_{3\alpha}^{\infty} \int_{3\alpha}^{z_2} \int_{2\alpha}^{z_4} \int_{0}^{z_3}         f_1(z_1) f_3(z_3) f_4(z_4) f_2(z_2)~\,dz_1\,dz_3\, dz_4\, dz_2 =\dfrac{1}{2}e^{-2\lambda \alpha}-\dfrac{1}{3}e^{-3\lambda \alpha}-\dfrac{1}{4}e^{-4\lambda \alpha} +\dfrac{1}{8}e^{-6\lambda \alpha} \nonumber\\
p_{13}=p(CADB|ABCD)=&\int_{3\alpha}^{\infty} \int_{3\alpha}^{z_2} \int_{2\alpha}^{z_4} \int_{2\alpha}^{z_1}    f_3(z_3) f_1(z_1) f_4(z_4) f_2(z_2)~\,dz_3\,dz_1\, dz_4\, dz_2 =\dfrac{1}{4}e^{-4\lambda \alpha}-\dfrac{1}{3}e^{-5\lambda \alpha}+\dfrac{1}{8}e^{-6\lambda \alpha}\nonumber\\
p_{14}=p(ADCB|ABCD)=&\int_{3\alpha}^{\infty} \int_{3\alpha}^{z_2} \int_{3\alpha}^{z_3} \int_{0}^{z_4}          f_1(z_1) f_4(z_4) f_3(z_3) f_2(z_2)~\,dz_1\,dz_4\, dz_3\, dz_2 =\dfrac{1}{6}e^{-3\lambda \alpha}-\dfrac{1}{8}e^{-6\lambda \alpha} \nonumber\\
p_{15}=p(DACB|ABCD)=&\int_{3\alpha}^{\infty} \int_{3\alpha}^{z_2} \int_{3\alpha}^{z_3} \int_{3\alpha}^{z_1}    f_4(z_4) f_1(z_1) f_3(z_3) f_2(z_2)~\,dz_4\,dz_1\, dz_3\, dz_2 =\dfrac{1}{24}e^{-6\lambda \alpha} \nonumber\\
p_{16}=p(CDAB|ABCD)=&\int_{3\alpha}^{\infty} \int_{3\alpha}^{z_2} \int_{3\alpha}^{z_1} \int_{2\alpha}^{z_4}    f_3(z_3) f_4(z_4) f_1(z_1) f_2(z_2)~\,dz_1\,dz_3\, dz_1\, dz_2 = \dfrac{1}{6}e^{-5\lambda \alpha}-\dfrac{1}{8}e^{-6\lambda \alpha}\nonumber\\
p_{17}=p(DCAB|ABCD)=&\int_{3\alpha}^{\infty} \int_{3\alpha}^{z_2} \int_{3\alpha}^{z_1} \int_{3\alpha}^{z_3}    f_4(z_4) f_3(z_3) f_1(z_1) f_2(z_2)~\,dz_3\,dz_4\, dz_1\, dz_2 = \dfrac{1}{24}e^{-6\lambda \alpha}  \nonumber\\
p_{18}=p(BCDA|ABCD)=&\int_{3\alpha}^{\infty} \int_{3\alpha}^{z_1} \int_{2\alpha}^{z_4} \int_{\alpha}^{z_3}    f_2(z_2) f_3(z_3) f_4(z_4) f_1(z_1)~\,dz_2\,dz_3\, dz_4\, dz_1 =\dfrac{1}{2}e^{-3\lambda \alpha}-\dfrac{7}{12}e^{-4\lambda \alpha}+\dfrac{1}{8}e^{-6\lambda \alpha}    \nonumber\\
p_{19}=p(CBDA|ABCD)=&\int_{3\alpha}^{\infty} \int_{3\alpha}^{z_1} \int_{2\alpha}^{z_4} \int_{2\alpha}^{z_2}   f_3(z_3) f_2(z_2) f_4(z_4) f_1(z_1)~\,dz_3\,dz_2\, dz_4\, dz_1 = \dfrac{1}{4}e^{-4\lambda \alpha}-\dfrac{1}{3}e^{-5\lambda \alpha}+ \dfrac{1}{8}e^{-6\lambda \alpha} \nonumber\\
p_{20}=p(BDCA|ABCD)=&\int_{3\alpha}^{\infty} \int_{3\alpha}^{z_1} \int_{3\alpha}^{z_3} \int_{\alpha}^{z_4}    f_2(z_2) f_4(z_4) f_3(z_3) f_1(z_1)~\,dz_2\,dz_4\, dz_3\, dz_1 =\dfrac{1}{6}e^{-4\lambda \alpha} -\dfrac{1}{8}e^{-6\lambda \alpha} \nonumber\\
p_{21}=p(DBCA|ABCD)=&\int_{3\alpha}^{\infty} \int_{3\alpha}^{z_1} \int_{3\alpha}^{z_3} \int_{3\alpha}^{z_2}   f_4(z_4) f_2(z_2) f_3(z_3) f_1(z_1)~\,dz_4\,dz_2\, dz_3\, dz_1 =\dfrac{1}{24}e^{-6\lambda \alpha}\nonumber\\
p_{22}=p(CDBA|ABCD)=&\int_{3\alpha}^{\infty} \int_{3\alpha}^{z_1} \int_{3\alpha}^{z_2} \int_{2\alpha}^{z_4}   f_3(z_3) f_4(z_4) f_2(z_2) f_1(z_1)~\,dz_3\,dz_4\, dz_2\, dz_1 = \dfrac{1}{6}e^{-5\lambda \alpha}-\dfrac{1}{8}e^{-6\lambda \alpha}   \nonumber\\
p_{23}=p(DCBA|ABCD)=&\int_{3\alpha}^{\infty} \int_{3\alpha}^{z_1} \int_{3\alpha}^{z_2} \int_{3\alpha}^{z_3}   f_4(z_4) f_3(z_3) f_2(z_2) f_1(z_1)~\,dz_4\,dz_3\, dz_2\, dz_1 = \dfrac{1}{24}e^{-6\lambda \alpha}~~~. \nonumber
\end{split}
\end{equation} 
}

\vspace{1cm}

\begin{figure*}
\centering
   \includegraphics[scale=.6]{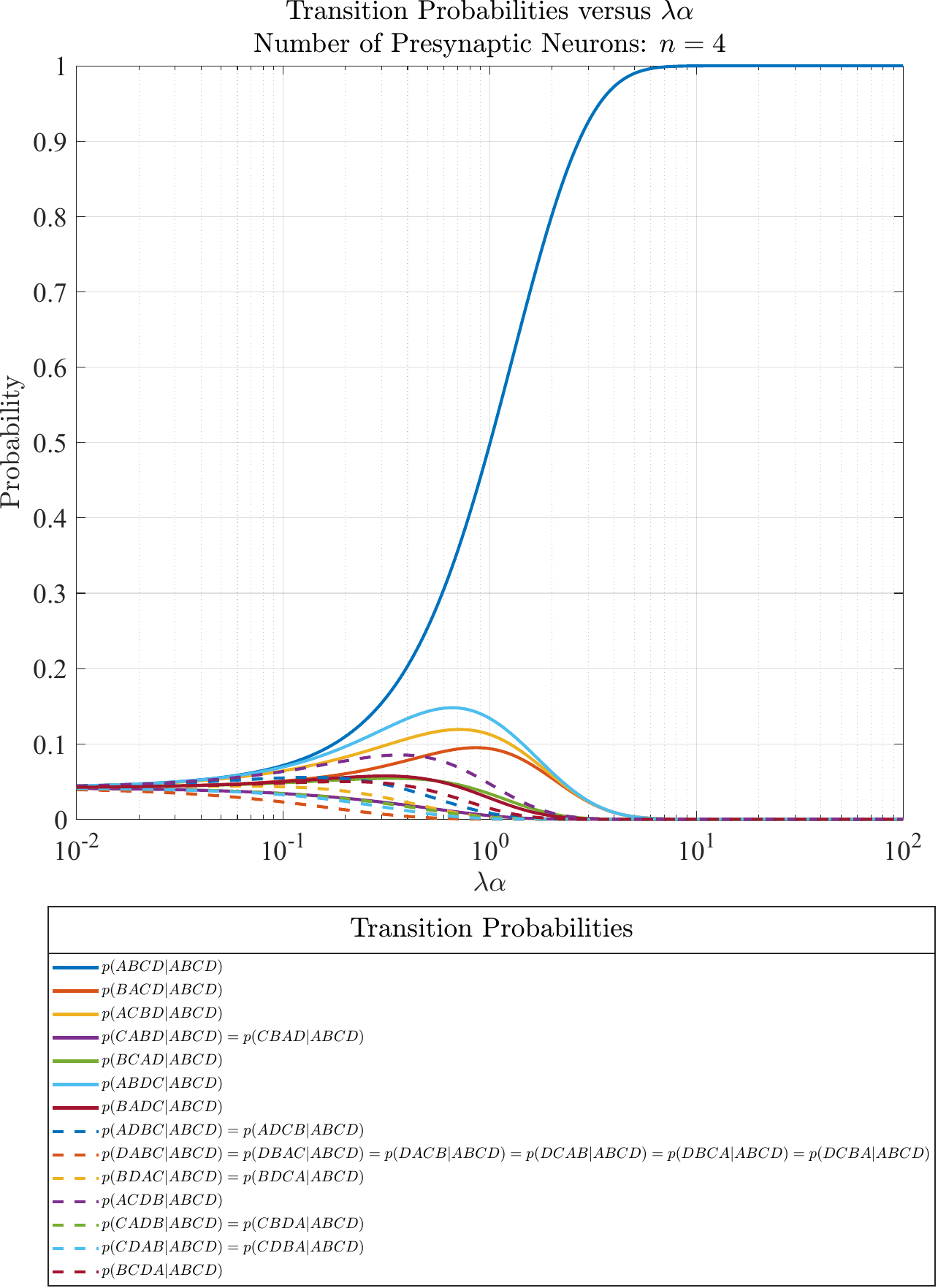}
   \caption{Transition probabilities for four presynaptic neurons ($n=4$) with exponential noise. Here $p(y|x)$ is the probability of sending sequence $x$ and receiving sequence $y$.}
   \label{fig:APPXPr4}
\end{figure*}

\noindent Fig.~\ref{fig:APPXPr4} shows the transition probabilities for four presynaptic neurons ($n=4$).

\clearpage
\newpage
\section{Transition Probabilities of Rank-Order Coding  with Gaussian Noise (Simulation)}\label{Appx2}

\begin{figure*}[tbh!]
\centering
   \includegraphics[scale=1.5]{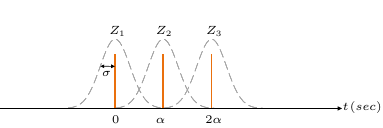}
   \caption{Spikes of rank-order coding with temporal noise (Guassian).}
   \label{fig:GN}
\end{figure*}

Consider Fig.~\ref{fig:GN}. In this section, we provide simulation results of transition probability when the temporal noise is Gaussian, $Z_i\sim \mathcal{N}\left((i-1)\alpha,\,\sigma^{2}\right)\,$. Here random variable $Z_i$ describes the random jitter caused by noise. The variance of the Gaussian noise is $\sigma^2$ and its mean is $(i-1)\alpha$, where index $i$ is for the $i^\text{th}$ number of presynaptic neuron, and $\alpha$ is the original spacing between successive spikes before noise is introduced.

\begin{figure*}[tbh!]
\centering
   \includegraphics[scale=.55]{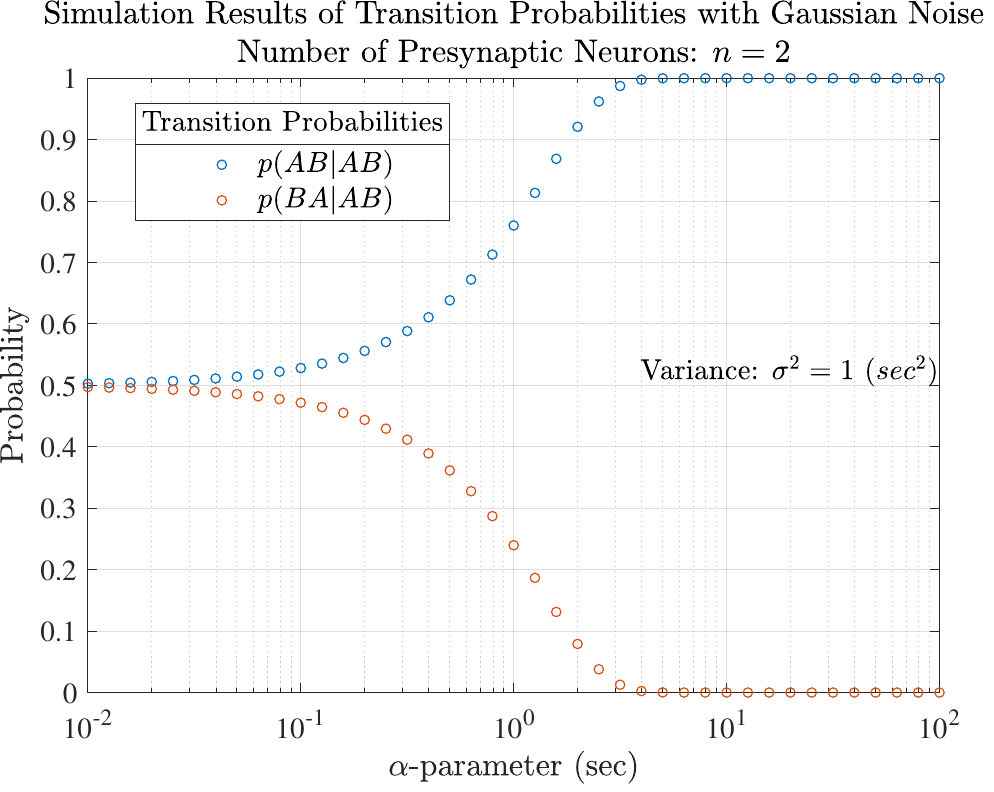}
   \caption{Simulation results of transition probabilities for two presynaptic neurons ($n=2$) with Gaussian noise. Here $p(y|x)$ is the probability of sending sequence $x$ and receiving sequence $y$. The variance of the Gaussian distribution, $\sigma^2$, is arbitrarily set to 1 (sec$^2$). Additionally, $\alpha$ is the original spacing between successive spikes before noise is introduced. The number of samples per point used in the simulation is $10^9$.}
   \label{fig:Pr2G}
\end{figure*}

\clearpage
\newpage
\section{Average Symbol Duration, $\bar{T}$}\label{Appx3}

Due to noise, the duration between the first ($Z_{(n)}$) and last spike ($Z_{(1)}$) of a rank-order coding symbol is a random variable $T= Z_{(n)} - Z_{(1)}$. Our goal in this section is to find the average value of $T$ for the set $\{ 2, 3, 4\}$ of presynaptic neurons. We first start by finding the \textit{pdfs} of $Z_{(1)}$ and  $Z_{(n)}$, and then their mean values. \\

\noindent Let $Z_i$ be a random variable and  $f_{Z_i}(z)$ its \textit{pdf}; here index $i$ indicates the $i^{\text{th}}$ presynaptic neuron. This \textit{pdf} describes the likelihood of observing a randomly jittered spike, by noise,  at time $z$ and is given by

\begin{equation} \label{eq:pdf}
f_{Z_i}(z)=
\begin{cases}
\lambda~e^{-\lambda \bigl (z-(i-1)\alpha\bigr)}&,~\text{if}~z\geq(i-1)\alpha\\
0&,~\text{otherwise}
\end{cases}
\end{equation}

\noindent here $\alpha$ is the spacing between successive spikes before noise is introduced, and $\lambda$ is the rate parameter of the exponential distribution of the noise. Herein random variables, $Z_i$, are independent.\\


\subsection{Two  Presynaptic Neurons}
\noindent\textbf{Minimum}. 
Let random variable $Z_{(1)}$ be the first-order statistic  defined as

\begin{equation}    
Z_{(1)}=\min\{Z_1,Z_2\}~~.
\end{equation}


\noindent The \textit{pdf} of $Z_{(1)}$ has three cases:\\

\noindent\textbf{Case 1: $ z< 0$}
\begin{equation} \label{eq1_1}
\begin{split}
Z_{(1)}=&\min\{Z_1,Z_2\}=\varnothing~\\[0.25cm]
\end{split}
\end{equation}
 because if $z < 0$, then $f_{Z_1}(z)=f_{Z_2}(z)=0$, (Eq.~\ref{eq:pdf}). Here, $\varnothing$ denotes the empty set.\\
 Therefore, 
\begin{equation}
f_{Z_{(1)}}(z)= 0~.
\end{equation}

\noindent\textbf{Case 2: $0\le z < \alpha$}

\begin{equation} \label{eq1_2}
\begin{split}
Z_{(1)}=&\min\{Z_1,Z_2\}=Z_1~~\\[0.25cm]
\end{split}
\end{equation}
 because in the range $0\le z < \alpha$, $f_{Z_2}(z)=0$, (Eq.~\ref{eq:pdf}).
 Therefore,
\begin{equation}
f_{Z_{(1)}}(z)= \lambda e^{-\lambda{z}}~.
\end{equation}

\vspace{0.5cm}

\noindent\textbf{Case 3: $\alpha\le z $}
\begin{equation}    
\begin{split}
Z_{(1)}=&\min\{Z_1,Z_2\}.
\end{split}
\end{equation}

The \textit{cdf} of the minimum is
\begin{equation} 
\begin{split}
F_{Z_{(1)}}(z)=&P\left(Z_{(1)}\le z\right)\\
=&1-P\left(Z_{(1)}> z\right)\\
=&1-P(Z_1>z,Z_2>z)\\
=&1-P(Z_1>z)(Z_2>z)~~~~\text{because of independence}\\
=&1-\Bigl[\Bigl(1-P(Z_1\le z)\Bigr)\Bigl(1-P(Z_2\le z)\Bigr)  \Bigr]\\
=&1-e^{-2\lambda (z-\frac{1}{2}\alpha)}~~.
\end{split}
\end{equation}

Accordingly, the \textit{pdf} of the minimum is

\begin{equation}
f_{Z_{(1)}}(z)=\frac{d}{dz}F_{Z_{(1)}}(z)=2\lambda e^{-2\lambda (z-\frac{1}{2}\alpha)}~~.
\end{equation}

\vspace{1cm}

Combining \textbf{Case 1} to \textbf{3}:
\begin{equation} \label{eq:Final_pdf2}
f_{Z_{(1)}}(z)=
\begin{cases}
0&,~\text{if}~z < 0\\[0.25cm]
\lambda e^{-\lambda{z}}&,~\text{if}~0\le z < \alpha\\[0.25cm]
2\lambda e^{-2\lambda (z-\frac{1}{2}\alpha)}&,~\text{if}~\alpha\le z ~~~.
\end{cases}
\end{equation}

\noindent Using Eq.~\ref{eq:Final_pdf2}, the mean value of  $Z_{(1)}$ is
\begin{equation} \label{eq:mean_Min2}
\mathbb{E}[Z_{(1)}]=\int_{-\infty}^{\infty} z f_{Z_{(1)}}(z) \,dz =\frac{1}{\lambda}-\frac{1}{2\lambda}e^{-\lambda \alpha}~~~\text{(sec/symbol)}.
\end{equation}

 \vspace{1cm}

\noindent\textbf{Maximum}. 
Let random variable $Z_{(2)}$ be the second-order statistic  defined as

\begin{equation}    
Z_{(2)}=\max\{Z_1,Z_2\}~.
\end{equation}

\noindent The \textit{pdf} of $Z_{(2)}$ has two cases:\\

\noindent\textbf{Case 1: $ z < \alpha$}
\begin{equation}    
Z_{(2)}=\max\{Z_1,Z_2\}=\varnothing~
\end{equation}
\noindent because for two neurons the maximum, that is the second-order statistic, needs to be greater or equal to  $\alpha$ (Eq.~\ref{eq:pdf}).
 \noindent Therefore,
\begin{equation} \label{eqm2_1}
\begin{split}
f_{Z_{(2)}}(z)=&~0~.
\end{split}
\end{equation}

\vspace{0.5cm}

\noindent\textbf{Case 2: $z \ge \alpha$}
\begin{equation}    
Z_{(2)}=\max\{Z_1,Z_2\}.
\end{equation}

The  \textit{cdf} of the maximum is
\begin{equation} 
\begin{split}
F_{Z_{(2)}}(z)=&P\left(Z_{(2)}\le z\right)\\
=&P(Z_1\le z,Z_2 \le z)\\
=&P(Z_1 \le z)(Z_2\le z)~~~~\text{because of independence}\\
=&1-e^{-\lambda z}-e^{-\lambda (z-\alpha)}+e^{-2\lambda (z-\frac{1}{2}\alpha)}~~.
\end{split}
\end{equation}

Accordingly, the \textit{pdf} of the maximum is

\begin{equation}
f_{Z_{(2)}}(z)=\frac{d}{dz}F_{Z_{(2)}}(z)=\lambda e^{-\lambda z}+\lambda e^{-\lambda (z-\alpha)}-2\lambda e^{-2\lambda (z-\frac{1}{2}\alpha)}~~.
\end{equation}

\vspace{1cm}

Combining \textbf{Case 1} and \textbf{2}:
\begin{equation} \label{eq:Final_pdf_Max2}
f_{Z_{(2)}}(z)=
\begin{cases}
0&,~\text{if}~z < \alpha\\[0.25cm]
\lambda e^{-\lambda z}+\lambda e^{-\lambda (z-\alpha)}-2\lambda e^{-2\lambda (z-\frac{1}{2}\alpha)}&,~\text{if}~\alpha\le z ~~~.
\end{cases}
\end{equation}

\noindent Using Eq.~\ref{eq:Final_pdf_Max2}, the mean value of  $Z_{(2)}$ is
\begin{equation} \label{eq:mean_Max2}
\mathbb{E}[Z_{(2)}]=\int_{-\infty}^{\infty} z f_{Z_{(2)}}(z) \,dz =\frac{\lambda\alpha+1}{\lambda}+\frac{1}{2\lambda}e^{-\lambda \alpha}~~~\text{(sec/symbol)}.
\end{equation} 

\noindent Using Eq.~\ref{eq:mean_Min2} and \ref{eq:mean_Max2}, the average symbol duration for two neurons is 
\begin{empheq}[box=\widefbox]{equation}\label{eq:T2}
\bar{T}=\mathbb{E}~[T]=\mathbb{E}~[Z_{(2)}-Z_{(1)}]=\mathbb{E}~[Z_{(2)}]-\mathbb{E}~[Z_{(1)}]= \alpha+\frac{1}{\lambda}e^{-\lambda \alpha}~~\text{(sec/symbol)}.
\end{empheq}

\begin{figure}
\centering
   \includegraphics[scale=.6]{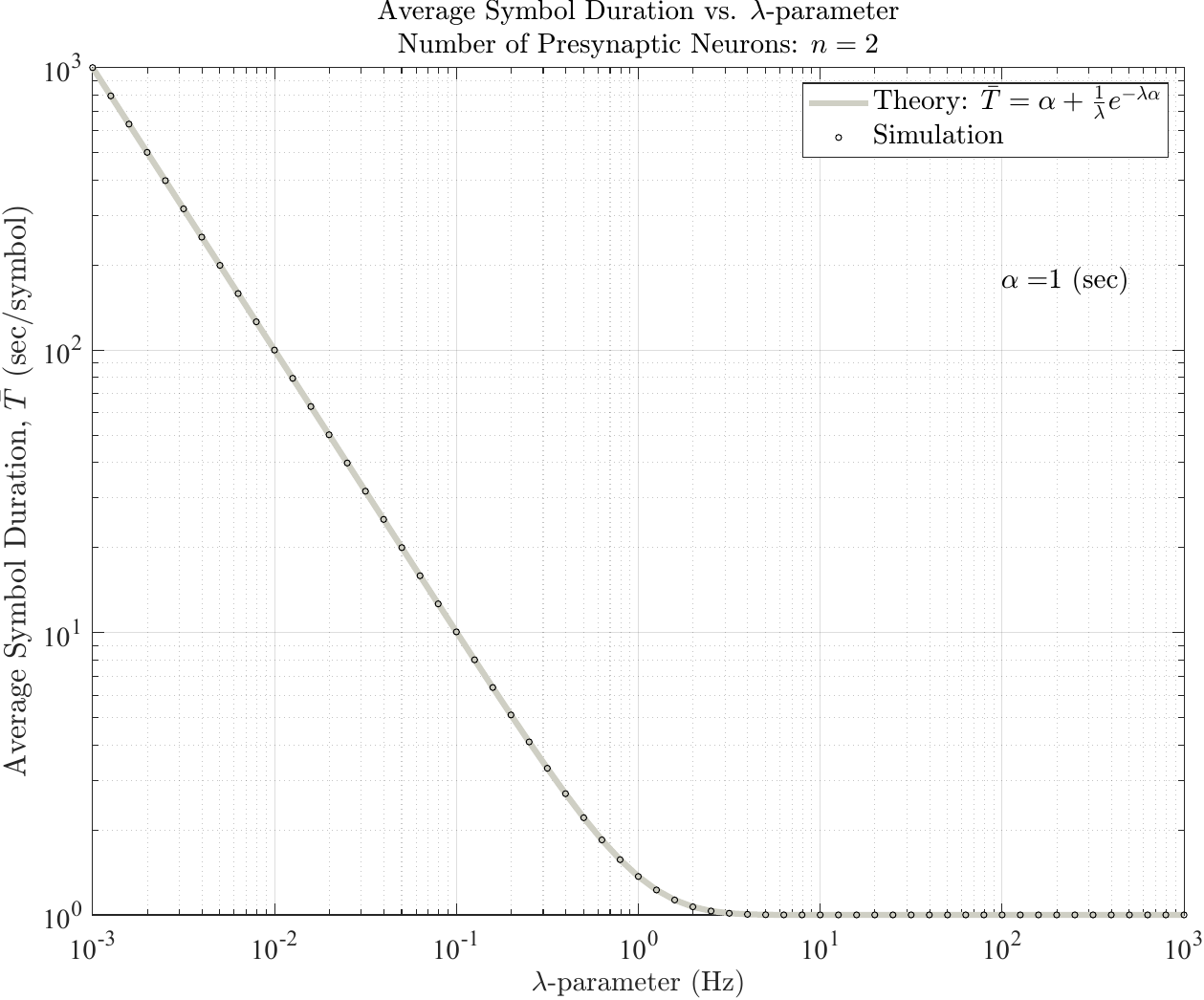}
   \caption{Theory (Eq.~\ref{eq:T2}) versus simulation for the average symbol duration, $\bar{T}$. The number of presynaptic neurons is $n=2$. Here $\alpha$ is arbitrarily set to 1 (sec), and  the number of samples per point used in the simulation is $10^9$.}
   \label{fig:S2} 
\end{figure}

Fig.~\ref{fig:S2} compares theory (Eq.~\ref{eq:T2}) with simulation. 

\newpage
\clearpage


\subsection{Three Presynaptic Neurons}
\noindent\textbf{Minimum}. 
Let random variable $Z_{(1)}$ be the first-order statistic  defined as

\begin{equation}    
Z_{(1)}=\min\{Z_1,Z_2,Z_3\}~.
\end{equation}


\noindent The \textit{pdf} of $Z_{(1)}$ has four cases:\\

\noindent\textbf{Case 1: $ z< 0$}
\begin{equation} \label{eq1_3}
\begin{split}
Z_{(1)}=&\min\{Z_1,Z_2,Z_3\}=\varnothing~~\\[0.25cm]
\end{split}
\end{equation}
 because if $ z < 0$, then $f_{Z_1}(z)=f_{Z_2}(z)=f_{Z_3}(z)=0$, (Eq.~\ref{eq:pdf}).
 Therefore,
\begin{equation}
f_{Z_{(1)}}(z)= 0~.
\end{equation}

\noindent\textbf{Case 2: $0\le z < \alpha$}

\begin{equation} \label{eq1_4}
\begin{split}
Z_{(1)}=&\min\{Z_1,Z_2,Z_3\}=Z_1~~\\[0.25cm]
\end{split}
\end{equation}
 because in the range $0\le z < \alpha$, $f_{Z_2}(z)=f_{Z_3}(z)=0$, (Eq.~\ref{eq:pdf}).
 Therefore,
\begin{equation}
f_{Z_{(1)}}(z)= \lambda e^{-\lambda{z}}~.
\end{equation}

\vspace{0.5cm}

\noindent\textbf{Case 3: $\alpha\le z <2\alpha$}
\begin{equation}    
\begin{split}
Z_{(1)}=&\min\{Z_1,Z_2,Z_3\}=\min\{Z_1,Z_2\}~
\end{split}
\end{equation}
because in the range $\alpha\le z <2\alpha$, $f_{Z_3}(z)=0$, (Eq.~\ref{eq:pdf}).

\vspace{0.5cm}
The \textit{cdf} of the minimum is
\begin{equation} 
\begin{split}
F_{Z_{(1)}}(z)=&P\left(Z_{(1)}\le z\right)\\
=&1-P\left(Z_{(1)}> z\right)\\
=&1-P(Z_1>z,Z_2>z)\\
=&1-P(Z_1>z)(Z_2>z)~~~~\text{because of independence}\\
=&1-\Bigl[\Bigl(1-P(Z_1\le z)\Bigr)\Bigl(1-P(Z_2\le z)\Bigr)  \Bigr]\\
=&1-e^{-2\lambda (z-\frac{1}{2}\alpha)}~~.
\end{split}
\end{equation}

Accordingly, the \textit{pdf} of the minimum is

\begin{equation}
f_{Z_{(1)}}(z)=\frac{d}{dz}F_{Z_{(1)}}(z)=2\lambda e^{-2\lambda (z-\frac{1}{2}\alpha)}~~.
\end{equation}

\noindent\textbf{Case 4: $2\alpha\le z $}
\begin{equation}    
\begin{split}
Z_{(1)}=&\min\{Z_1,Z_2,Z_3\}.
\end{split}
\end{equation}

The \textit{cdf} of the minimum is
\begin{equation} 
\begin{split}
F_{Z_{(1)}}(z)=&P\left(Z_{(1)}\le z\right)\\
=&1-P\left(Z_{(1)}> z\right)\\
=&1-P(Z_1>z,Z_2>z,Z_3>z)\\
=&1-P(Z_1>z)(Z_2>z)(Z_3>z)~~~~\text{because of independence}\\
=&1-\Bigl[\Bigl(1-P(Z_1\le z)\Bigr)\Bigl(1-P(Z_2\le z)\Bigr)  \Bigl(1-P(Z_3\le z)\Bigr) \Bigr]\\
=&1-e^{-3\lambda (z-\alpha)}~~.
\end{split}
\end{equation}

Accordingly, the \textit{pdf} of the minimum is

\begin{equation}
f_{Z_{(1)}}(z)=\frac{d}{dz}F_{Z_{(1)}}(z)=3\lambda e^{-3\lambda (z-\alpha)}~~.
\end{equation}

\vspace{1cm}

Combining \textbf{Case 1} to \textbf{4}:
\begin{equation} \label{eq:Final_pdf3}
f_{Z_{(1)}}(z)=
\begin{cases}
0&,~\text{if}~z < 0\\[0.25cm]
\lambda e^{-\lambda{z}}&,~\text{if}~0\le z < \alpha\\[0.25cm]
2\lambda e^{-2\lambda (z-\frac{1}{2}\alpha)}&,~\text{if}~\alpha\le z < 2\alpha \\[0.25cm]
3\lambda e^{-3\lambda (z-\alpha)}&,~\text{if}~2\alpha\le z ~~.

\end{cases}
\end{equation}

\noindent Using Eq.~\ref{eq:Final_pdf3}, the mean value of  $Z_{(1)}$ is
\begin{equation} \label{eq:mean_Min}
\mathbb{E}[Z_{(1)}]=\int_{-\infty}^{\infty} z f_{Z_{(1)}}(z) \,dz =\frac{1}{\lambda}-\frac{1}{2\lambda}e^{-\lambda \alpha} -\frac{1}{6\lambda}e^{-3\lambda \alpha}~~~\text{(sec/symbol)}.
\end{equation}

 \vspace{1cm}

\noindent\textbf{Maximum}. 
Let random variable $Z_{(3)}$ be the third-order statistic  defined as

\begin{equation}    
Z_{(3)}=\max\{Z_1,Z_2,Z_3\}~.
\end{equation}

\noindent The \textit{pdf} of $Z_{(3)}$ has two cases:\\

\noindent\textbf{Case 1: $ z < 2\alpha$}
\begin{equation}    
Z_{(3)}=\max\{Z_1,Z_2,Z_3\}=\varnothing~
\end{equation}
\noindent because for three neurons the maximum, that is the third-order statistic, needs to be greateror equal to  $2\alpha$ (Eq.~\ref{eq:pdf}).
 \noindent Therefore,
\begin{equation} \label{eqm2_2}
\begin{split}
f_{Z_{(3)}}(z)=&~0~.
\end{split}
\end{equation}

\vspace{0.5cm}

\noindent\textbf{Case 2: $z \ge 2\alpha$}
\begin{equation}    
Z_{(3)}=\max\{Z_1,Z_2,Z_3\}.
\end{equation}

The  \textit{cdf} of the maximum is
\begin{equation} 
\begin{split}
F_{Z_{(3)}}(z)=&P\left(Z_{(3)}\le z\right)\\
=&P(Z_1\le z,Z_2 \le z,Z_3 \le z)\\
=&P(Z_1 \le z)(Z_2\le z)(Z_3\le z)~~~~\text{because of independence}\\
=&1-e^{-\lambda (z-\alpha)}-e^{-\lambda z}+e^{-\lambda (2z-\alpha)}-e^{-\lambda (z-2\alpha)}+e^{-\lambda (2z-3\alpha)}+e^{-2\lambda (z-\alpha)}-e^{-3\lambda (z-\alpha)}~~.
\end{split}
\end{equation}

Accordingly, the \textit{pdf} of the maximum is

\begin{equation}
f_{Z_{(3)}}(z)=\frac{d}{dz}F_{Z_{(3)}}(z)=\lambda e^{-\lambda (z-\alpha)}+\lambda e^{-\lambda z}-2\lambda e^{-\lambda (2z-\alpha)}+\lambda e^{-\lambda (z-2\alpha)}-2\lambda e^{-\lambda (2z-3\alpha)}-2\lambda e^{-2\lambda (z-\alpha)}+3\lambda e^{-3\lambda (z-\alpha)}.
\end{equation}

\vspace{0.5cm}

\noindent Combining \textbf{Case 1} and \textbf{2}:
\begin{equation} \label{eq:Final_pdf_Max3}
f_{Z_{(3)}}(z)=
\begin{cases}
0&,~\text{if}~z < 2\alpha\\[0.25cm]
\lambda e^{-\lambda (z-\alpha)}+\lambda e^{-\lambda z}-2\lambda e^{-\lambda (2z-\alpha)}+\lambda e^{-\lambda (z-2\alpha)}-2\lambda e^{-\lambda (2z-3\alpha)}-2\lambda e^{-2\lambda (z-\alpha)}+3\lambda e^{-3\lambda (z-\alpha)}&,~\text{if}~2\alpha\le z
\end{cases}
\end{equation}

\noindent Using Eq.~\ref{eq:Final_pdf_Max3}, the mean value of  $Z_{(3)}$ is
\begin{equation} \label{eq:mean_Max}
\mathbb{E}[Z_{(3)}]=\int_{-\infty}^{\infty} z f_{Z_{(3)}}(z) \,dz =\frac{2\lambda\alpha+1}{\lambda}+\frac{1}{2\lambda}e^{-\lambda \alpha}+\frac{1}{2\lambda}e^{-2\lambda \alpha}-\frac{1}{6\lambda}e^{-3\lambda \alpha}~~~\text{(sec/symbol)}.
\end{equation} 

\noindent Using Eq.~\ref{eq:mean_Min} and \ref{eq:mean_Max}, the average symbol duration for three neurons is 
\begin{empheq}[box=\widefbox]{equation}\label{eq:T33}
\bar{T}=\mathbb{E}~[T]=\mathbb{E}~[Z_{(3)}-Z_{(1)}]=\mathbb{E}~[Z_{(3)}]-\mathbb{E}~[Z_{(1)}]= 2\alpha+\frac{1}{\lambda}e^{-\lambda \alpha}+\frac{1}{2\lambda}e^{-2\lambda \alpha}~~\text{(sec/symbol)}.
\end{empheq}

\begin{figure}
\centering
   \includegraphics[scale=.6]{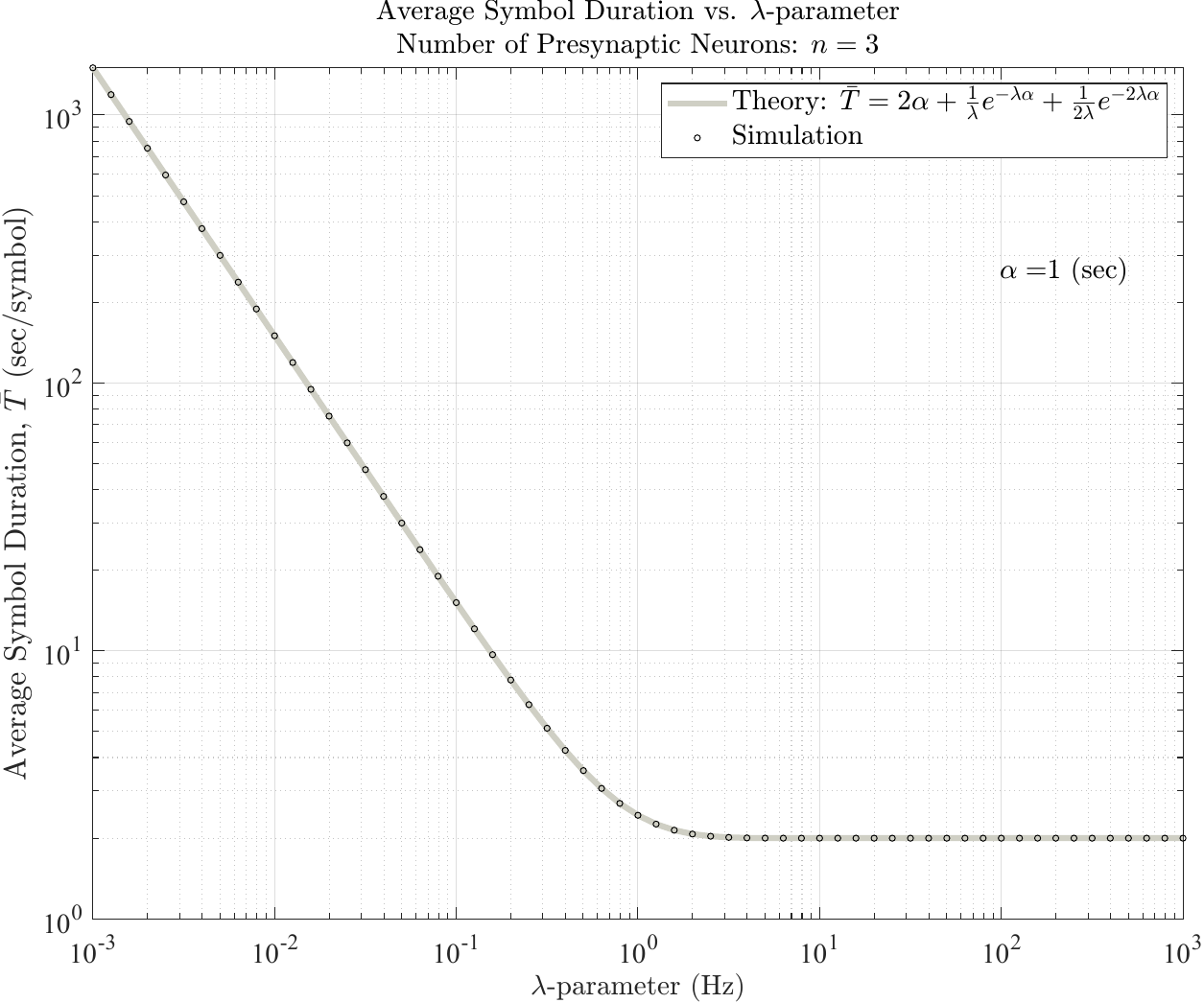}
   \caption{Theory (Eq.~\ref{eq:T33})  versus simulation for the average symbol duration, $\bar{T}$. The number of presynaptic neurons is $n=3$. Here $\alpha$ is arbitrarily set to 1 (sec), and  the number of samples per point used in the simulation is $10^9$.}
   \label{fig:S3} 
\end{figure}

Fig.~\ref{fig:S3} compares theory (Eq.~\ref{eq:T33}) with simulation. 

\newpage
\clearpage

\subsection{Four Presynaptic Neurons}
\noindent\textbf{Minimum}. 
Let random variable $Z_{(1)}$ be the first-order statistic  defined as

\begin{equation}    
Z_{(1)}=\min\{Z_1,Z_2,Z_3,Z_4\}~.
\end{equation}


\noindent The \textit{pdf} of $Z_{(1)}$ has five cases:\\

\noindent\textbf{Case 1: $ z< 0$}
\begin{equation} \label{eq1_5}
\begin{split}
Z_{(1)}=&\min\{Z_1,Z_2,Z_3,Z_4\}=\varnothing~~\\[0.25cm]
\end{split}
\end{equation}
 \noindent because if $ z < 0$, then $f_{Z_1}(z)=f_{Z_2}(z)=f_{Z_3}(z)=f_{Z_4}(z)=0$, (Eq.~\ref{eq:pdf}).
  \noindent Therefore,
\begin{equation}
f_{Z_{(1)}}(z)= 0~.
\end{equation}

\noindent\textbf{Case 2: $0\le z < \alpha$}

\begin{equation} \label{eq1}
\begin{split}
Z_{(1)}=&\min\{Z_1,Z_2,Z_3,Z_4\}=Z_1~~\\[0.25cm]
\end{split}
\end{equation}
 because in the range $0\le z < \alpha$, $f_{Z_2}(z)=f_{Z_3}(z)=f_{Z_4}(z)=0$, (Eq.~\ref{eq:pdf}).
 Therefore,
\begin{equation}
f_{Z_{(1)}}(z)= \lambda e^{-\lambda{z}}~.
\end{equation}

\vspace{0.5cm}

\noindent\textbf{Case 3: $\alpha\le z <2\alpha$}
\begin{equation}    
\begin{split}
Z_{(1)}=&\min\{Z_1,Z_2,Z_3,Z_4\}=\min\{Z_1,Z_2\}~
\end{split}
\end{equation}
because in the range $\alpha\le z <2\alpha$, $f_{Z_3}(z)=f_{Z_4}(z)=0$ .

\vspace{0.5cm}
The \textit{cdf} of the minimum is
\begin{equation} 
\begin{split}
F_{Z_{(1)}}(z)=&P\left(Z_{(1)}\le z\right)\\
=&1-P\left(Z_{(1)}> z\right)\\
=&1-P(Z_1>z,Z_2>z)\\
=&1-P(Z_1>z)(Z_2>z)~~~~\text{because of independence}\\
=&1-\Bigl[\Bigl(1-P(Z_1\le z)\Bigr)\Bigl(1-P(Z_2\le z)\Bigr)  \Bigr]\\
=&1-e^{-2\lambda (z-\frac{1}{2}\alpha)}~~.
\end{split}
\end{equation}

Accordingly, the \textit{pdf} of the minimum is

\begin{equation}
f_{Z_{(1)}}(z)=\frac{d}{dz}F_{Z_{(1)}}(z)=2\lambda e^{-2\lambda (z-\frac{1}{2}\alpha)}~~~.
\end{equation}

\noindent\textbf{Case 4: $2\alpha\le z <3\alpha$}
\begin{equation}    
\begin{split}
Z_{(1)}=&\min\{Z_1,Z_2,Z_3,Z_4\}= \min\{Z_1,Z_2,Z_3\}
\end{split}
\end{equation}
because in the range $2\alpha\le z <3\alpha$, $f_{Z_4}(z)=0$.\\
The \textit{cdf} of the minimum is
\begin{equation} 
\begin{split}
F_{Z_{(1)}}(z)=&P\left(Z_{(1)}\le z\right)\\
=&1-P\left(Z_{(1)}> z\right)\\
=&1-P(Z_1>z,Z_2>z,Z_3>z)\\
=&1-P(Z_1>z)(Z_2>z)(Z_3>z)~~~~\text{because of independence}\\
=&1-\Bigl[\Bigl(1-P(Z_1\le z)\Bigr)\Bigl(1-P(Z_2\le z)\Bigr)  \Bigl(1-P(Z_3\le z)\Bigr) \Bigr]\\
=&1-e^{-3\lambda (z-\alpha)}~~.
\end{split}
\end{equation}

Accordingly, the \textit{pdf} of the minimum is

\begin{equation}
f_{Z_{(1)}}(z)=\frac{d}{dz}F_{Z_{(1)}}(z)=3\lambda e^{-3\lambda (z-\alpha)}~~.
\end{equation}

\noindent\textbf{Case 5: $3\alpha\le z $}
\begin{equation}    
\begin{split}
Z_{(1)}=&\min\{Z_1,Z_2,Z_3,Z_4\}.
\end{split}
\end{equation}

The \textit{cdf} of the minimum is
\begin{equation} 
\begin{split}
F_{Z_{(1)}}(z)=&P\left(Z_{(1)}\le z\right)\\
=&1-P\left(Z_{(1)}> z\right)\\
=&1-P(Z_1>z,Z_2>z,Z_3>z,Z_4>z)\\
=&1-P(Z_1>z)(Z_2>z)(Z_3>z) (Z_4>z)~~~~\text{because of independence}\\
=&1-\Bigl[\Bigl(1-P(Z_1\le z)\Bigr)\Bigl(1-P(Z_2\le z)\Bigr)  \Bigl(1-P(Z_3\le z)\Bigr)  \Bigl(1-P(Z_4\le z)\Bigr)\Bigr]\\
=&1-e^{-4\lambda (z-\frac{3}{2}\alpha)}~~.
\end{split}
\end{equation}

Accordingly, the \textit{pdf} of the minimum is

\begin{equation}
f_{Z_{(1)}}(z)=\frac{d}{dz}F_{Z_{(1)}}(z)=4\lambda e^{-4\lambda (z-\frac{3}{2}\alpha)}~~.
\end{equation}

\vspace{1cm}

Combining \textbf{Case 1} to \textbf{5}:
\begin{equation} \label{eq:Final_pdf4}
f_{Z_{(1)}}(z)=
\begin{cases}
0&,~\text{if}~z < 0\\[0.25cm]
\lambda e^{-\lambda{z}}&,~\text{if}~0\le z < \alpha\\[0.25cm]
2\lambda e^{-2\lambda (z-\frac{1}{2}\alpha)}&,~\text{if}~\alpha\le z < 2\alpha \\[0.25cm]
3\lambda e^{-3\lambda (z-\alpha)}&,~\text{if}~2\alpha\le z<3\alpha\\[0.25cm] 
4\lambda e^{-4\lambda (z-\frac{3}{2}\alpha)}&,~\text{if}~3\alpha\le z~~.
\end{cases}
\end{equation}

\noindent Using Eq.~\ref{eq:Final_pdf4}, the mean value of  $Z_{(1)}$ is
\begin{equation} \label{eq:mean_Min4}
\mathbb{E}[Z_{(1)}]=\int_{-\infty}^{\infty} z f_{Z_{(1)}}(z) \,dz =\frac{1}{\lambda}-\frac{1}{2\lambda}e^{-\lambda \alpha} -\frac{1}{6\lambda}e^{-3\lambda \alpha}-\frac{1}{12\lambda}e^{-6\lambda \alpha}~~~\text{(sec/symbol)}.
\end{equation}

 \vspace{1cm}

\noindent\textbf{Maximum}. 
Let random variable $Z_{(4)}$ be the fourth-order statistic  defined as

\begin{equation}    
Z_{(4)}=\max\{Z_1,Z_2,Z_3,Z_4\}~.
\end{equation}

\noindent The \textit{pdf} of $Z_{(4)}$ has two cases:\\

\noindent\textbf{Case 1: $ z < 3\alpha$}
\begin{equation}    
Z_{(4)}=\max\{Z_1,Z_2,Z_3,Z_4\}=\varnothing~
\end{equation}
\noindent because for four neurons the maximum, that is the fourth-order statistic, needs to be greater or equal to  $3\alpha$ (Eq.~\ref{eq:pdf}).
 \noindent Therefore,
\begin{equation} \label{eqm2_3}
\begin{split}
f_{Z_{(4)}}(z)=&~0~~.
\end{split}
\end{equation}

\vspace{0.5cm}

\noindent\textbf{Case 2: $z \ge 3\alpha$}
\begin{equation}    
Z_{(4)}=\max\{Z_1,Z_2,Z_3,Z_4\}.
\end{equation}

The  \textit{cdf} of the maximum is

\begin{align}\label{eq:4}
F_{Z_{(4)}}(z)=&  P\left(Z_{(4)}\le z\right) \nonumber \\
=&P(Z_1\le z,Z_2 \le z,Z_3 \le z,Z_4 \le z) \nonumber \\
=&P(Z_1 \le z)(Z_2\le z)(Z_3\le z)(Z_4\le z)~~~~\text{because of independence}\nonumber \\
=&1-e^{-\lambda (z-\alpha)}-e^{-\lambda z}+e^{-\lambda (2z-\alpha)}-e^{-\lambda (z-2\alpha)}+2e^{-\lambda (2z-3\alpha)}+e^{-2\lambda (z-\alpha)}-e^{-3\lambda (z-\alpha)}\nonumber \\
&\qquad +e^{-2\lambda (z-2\alpha)}-e^{-\lambda (3z-4\alpha)}+e^{-\lambda (2z-5\alpha)}-e^{-3\lambda (z-2\alpha)}-e^{-\lambda (3z-5\alpha)}+e^{-2\lambda (2z-3\alpha)}-e^{-\lambda (z-3\alpha)}.
\end{align}

Accordingly, the \textit{pdf} of the maximum is

\begin{align}\label{eq:pdf4}
f_{Z_{(4)}}(z)=\frac{d}{dz}F_{Z_{(4)}}(z)=&\lambda e^{-\lambda (z-\alpha)}+\lambda e^{-\lambda z}-2\lambda e^{-\lambda (2z-\alpha)}+\lambda e^{-\lambda (z-2\alpha)}-4\lambda e^{-\lambda (2z-3\alpha)} \nonumber \\
&\qquad -2\lambda e^{-2\lambda (z-\alpha)}+3\lambda e^{-3\lambda (z-\alpha)} -2\lambda e^{-2\lambda (z-2\alpha)} +3\lambda e^{-\lambda (3z-4\alpha)}\nonumber \\
&\qquad -2 \lambda e^{-\lambda (2z-5\alpha)}+3\lambda e^{-3\lambda (z-2\alpha)}+3\lambda e^{-\lambda (3z-5\alpha)}-4\lambda e^{-2\lambda (2z-3\alpha)}+\lambda e^{-\lambda (z-3\alpha)}~~~. \nonumber \\
\end{align}

\vspace{0.5cm}

\noindent Combining \textbf{Case 1} and \textbf{2}:

\begin{equation} \label{eq:Final_pdf_Max4}
f_{Z_{(4)}}(z)=
\begin{cases}
0&,~\text{if}~z < 3\alpha\\[1.5cm]
\lambda e^{-\lambda (z-\alpha)}+\lambda e^{-\lambda z}-2\lambda e^{-\lambda (2z-\alpha)}+\lambda e^{-\lambda (z-2\alpha)}-4\lambda e^{-\lambda (2z-3\alpha)}&,~\text{if}~3\alpha\le z \\[0.25cm]
\qquad -2\lambda e^{-2\lambda (z-\alpha)}+3\lambda e^{-3\lambda (z-\alpha)} -2\lambda e^{-2\lambda (z-2\alpha)} +3\lambda e^{-\lambda (3z-4\alpha)}&\\[0.25cm]
\qquad  -2 \lambda e^{-\lambda (2z-5\alpha)}+3\lambda e^{-3\lambda (z-2\alpha)}+3\lambda e^{-\lambda (3z-5\alpha)}-4\lambda e^{-2\lambda (2z-3\alpha)}+\lambda e^{-\lambda (z-3\alpha)}~~.&
\end{cases}
\end{equation}

\noindent Using Eq.~\ref{eq:Final_pdf_Max4}, the mean value of  $Z_{(4)}$ is
\begin{equation} \label{eq:mean_Max4}
\mathbb{E}[Z_{(4)}]=\int_{-\infty}^{\infty} z f_{Z_{(4)}}(z) \,dz =\frac{3\lambda\alpha+1}{\lambda}+\frac{1}{2\lambda}e^{-\lambda \alpha}+\frac{1}{2\lambda}e^{-2\lambda \alpha}+\frac{1}{3\lambda}e^{-3\lambda \alpha}-\frac{1}{6\lambda}e^{-4\lambda \alpha}-\frac{1}{6\lambda}e^{-5\lambda \alpha}+\frac{1}{12\lambda}e^{-6\lambda \alpha}~~\text{(sec/symbol)}.
\end{equation} 

\noindent Using Eq.~\ref{eq:mean_Min4} and \ref{eq:mean_Max4}, the average symbol duration for four neurons is 
\begin{empheq}[box=\widefbox]{align}\label{eq:T44}
\bar{T}=\mathbb{E}~[T]=\mathbb{E}~[Z_{(4)}-Z_{(1)}]=\mathbb{E}~[Z_{(4)}]-\mathbb{E}~[Z_{(1)}]= 3\alpha+\frac{1}{\lambda}e^{-\lambda \alpha}+\frac{1}{2\lambda}e^{-2\lambda \alpha}+\frac{1}{2\lambda}e^{-3\lambda \alpha} \nonumber \\
-\frac{1}{6\lambda}e^{-4\lambda \alpha}-\frac{1}{6\lambda}e^{-5\lambda \alpha}+\frac{1}{6\lambda}e^{-6\lambda \alpha}~~~\text{(sec/symbol)}.
\end{empheq}

\begin{figure}
\centering
   \includegraphics[scale=.6]{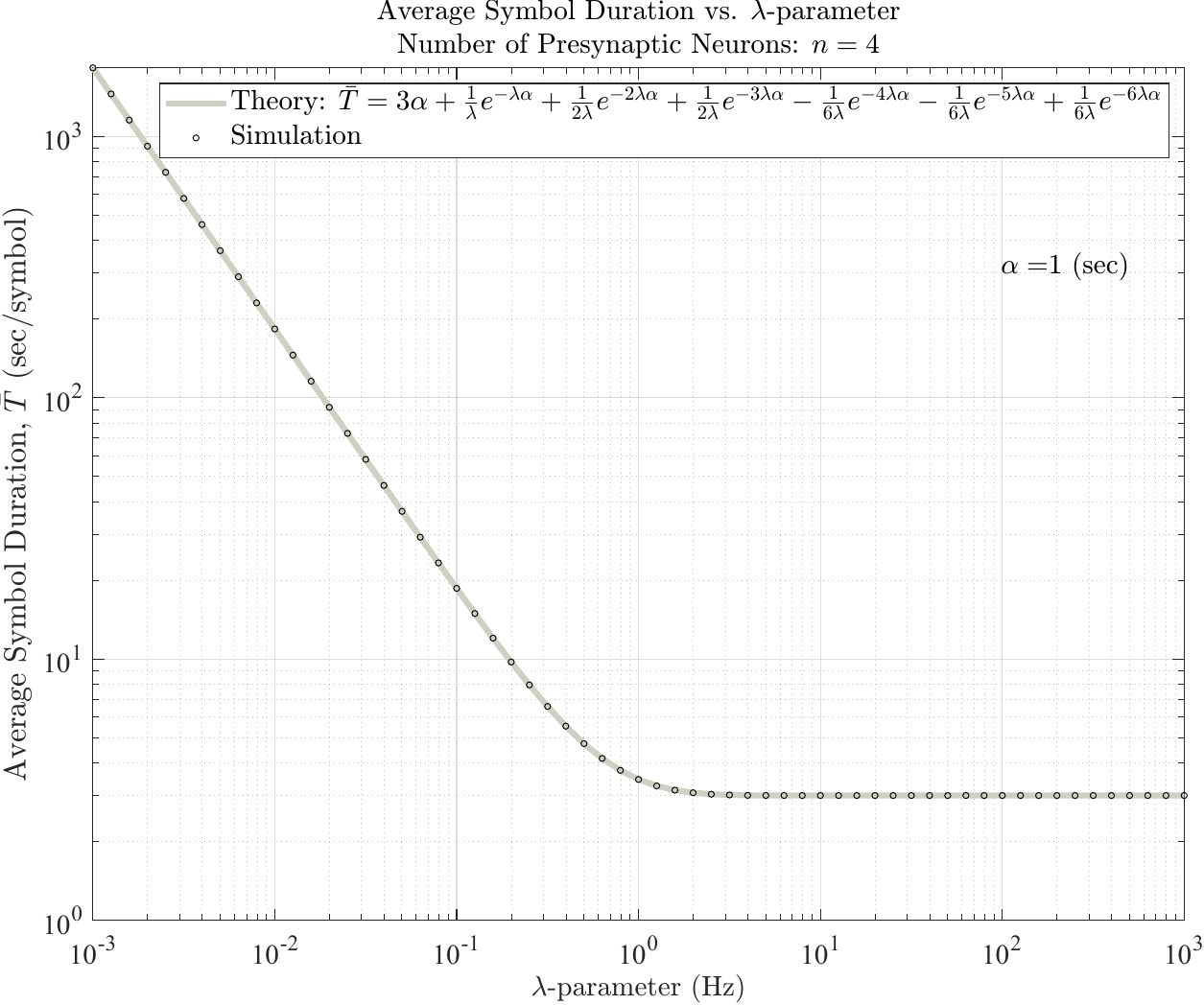}
   \caption{Theory (Eq.~\ref{eq:T44})  versus simulation for the average symbol duration, $\bar{T}$. The number of presynaptic neurons is $n=4$. Here $\alpha$ is arbitrarily set to 1 (sec), and  the number of samples per point used in the simulation is $10^9$.}
   \label{fig:S4} 
\end{figure}

Fig.~\ref{fig:S4} compares theory (Eq.~\ref{eq:T44}) with simulation. 

\end{document}